\pdfoutput=1
\documentclass[11pt]{article}

\usepackage[]{acl}

\usepackage{times}
\usepackage{amsmath}
\usepackage{amssymb}
\usepackage{graphicx}
\usepackage{caption}
\usepackage{subcaption}
\usepackage{xcolor}
\usepackage{float}
\usepackage{booktabs}
\usepackage{listings}
\usepackage{natbib}
\usepackage{latexsym}
\usepackage[textsize=scriptsize]{todonotes}
\usepackage{algorithm}
\usepackage{algpseudocode}
\usepackage{algorithmicx}
\usepackage{hyperref}
\usepackage{xr}
\usepackage{bbm}
\usepackage{enumitem}
\usepackage{listings}
\usepackage{multirow}
\usepackage{xcolor}
\usepackage[normalem]{ulem}
\usepackage{spverbatim}
\usepackage{tabularx}

\usepackage[T1]{fontenc}
\usepackage[utf8]{inputenc}

\usepackage{microtype}

\usepackage[textsize=scriptsize]{todonotes}

\newcommand{\circled}[1]{\raisebox{.5pt}{\textcircled{\raisebox{-.9pt} {#1}}}}

\usepackage{inconsolata}

\title{
Complex Claim Verification with Evidence Retrieved in the Wild
}

\author{Jifan Chen \quad\quad Grace Kim \quad\quad Aniruddh Sriram  \quad\quad Greg Durrett \quad\quad Eunsol Choi\\
  Department of Computer Science \\
  The University of Texas at Austin \\
  \texttt{jf\_chen@utexas.edu}}

\begin{document}
\maketitle
\begin{abstract}
Retrieving evidence to support or refute claims is a core part of automatic fact-checking. Prior work makes simplifying assumptions in retrieval that depart from real-world use cases: either no access to evidence, access to evidence curated by a human fact-checker, or access to evidence published after a claim was made. 
In this work, we present the first realistic pipeline to check real-world claims by retrieving raw evidence from the web. We restrict our retriever to only search documents available prior to the claim's making, modeling the realistic scenario of emerging claims.
Our pipeline includes five components: claim decomposition, raw document retrieval, fine-grained evidence retrieval, claim-focused summarization, and veracity judgment.  
We conduct experiments on complex political claims in the \textsc{ClaimDecomp} dataset and show that the aggregated evidence produced by our pipeline improves veracity judgments. Human evaluation finds the evidence summary produced by our system is reliable (it does not hallucinate information) and relevant to answering key questions about a claim, suggesting that it can assist fact-checkers even when it does not reflect a complete evidence set.\footnote{Code and data available at \url{https://github.com/jifan-chen/Fact-checking-via-Raw-Evidence}}
\end{abstract}
% (using GPT-3)

\section{Introduction}

\begin{figure}[t!]
  \centering
  \includegraphics[width=0.49\textwidth,trim=5mm 0 0 0]{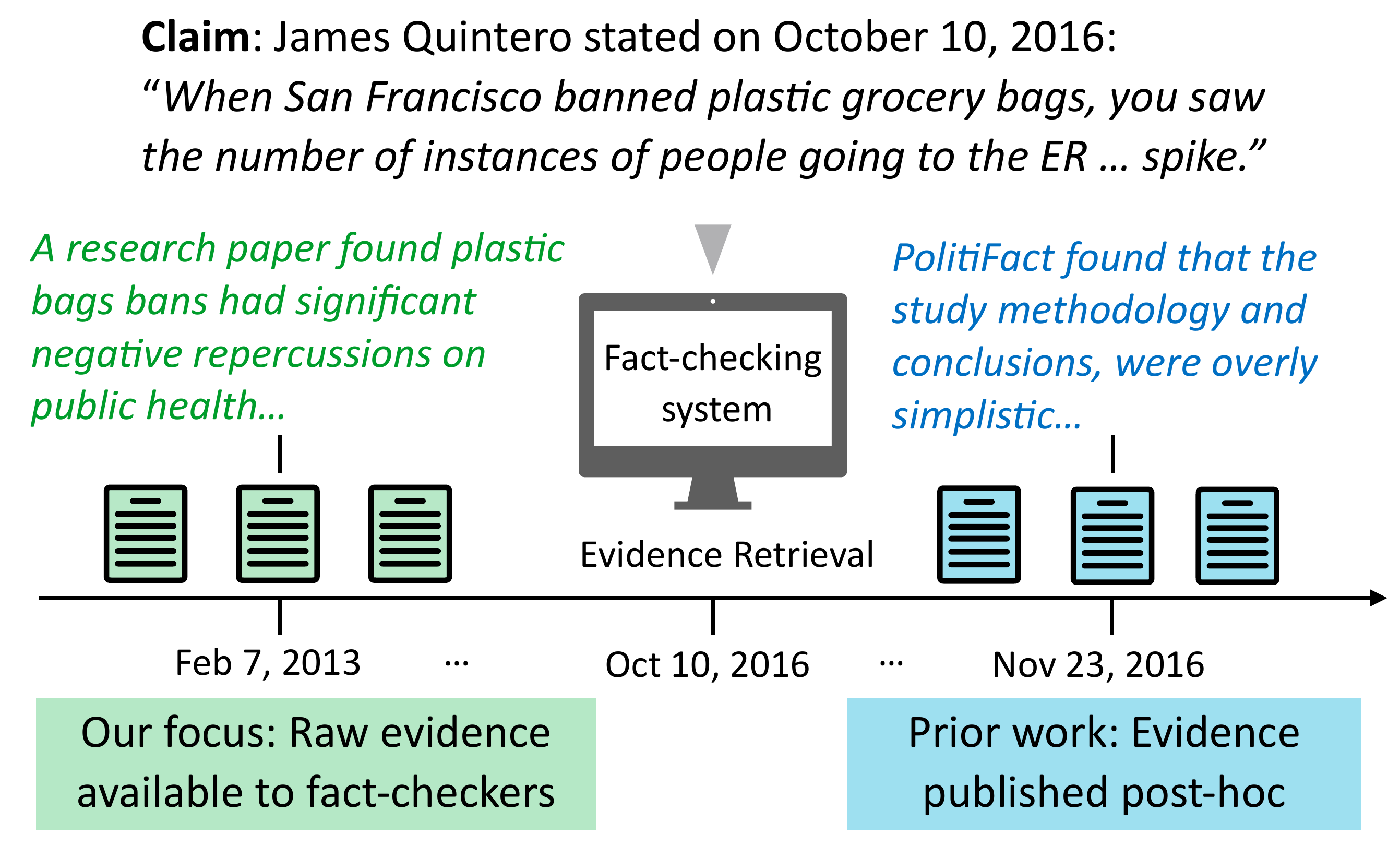}
  \caption{Our fact-checking setting addresses realistic claims using evidence retrieved prior to when the claim was made.}\vspace{-0.5em}
  \label{fig:introduction}
\end{figure}

To combat the rise of misinformation, the NLP community has developed automatic fact-checking tools.
%With the rise of misinformation on the web in recent years comes a greater need for fact-checking. Websites like PolitiFact\footnote{\url{www.politifact.com}} and Snopes\footnote{\url{www.snopes.com}} hire professional fact-checkers to manually check the claims, which involves identifying the claim, gathering evidence, verifying the claim and its reasoning chain. Due to the time-consuming nature of manual fact-checking, researchers have proposed numerous approaches for automating the process in recent years.
However, these automated systems are not ready for wide adoption at real fact-checking organizations. 
%Many studies have focused on crowd-authored claims~\cite{thorne-etal-2018-fever, jiang-etal-2020-hover, schuster-etal-2021-get, aly-etal-2021-fact}, which do not accurately represent the complexities of claims that fact-checkers deal with. 
Prior work handling real claims either relies on access to a document set which contains the ``gold'' evidence~\cite{ferreira2016emergent, alhindi-etal-2018-evidence, hanselowski-etal-2019-richly, atanasova-etal-2020-generating-fact} or conducts unconstrained retrieval~\cite{augenstein-etal-2019-multifc}, which may retrieve articles written by fact-checkers about the claim (example in Figure~\ref{fig:introduction}). %Prior work has not implemented a system to retrieve evidence in realistic settings.

\begin{figure*}[t!]
  \centering
  \includegraphics[width=\textwidth]{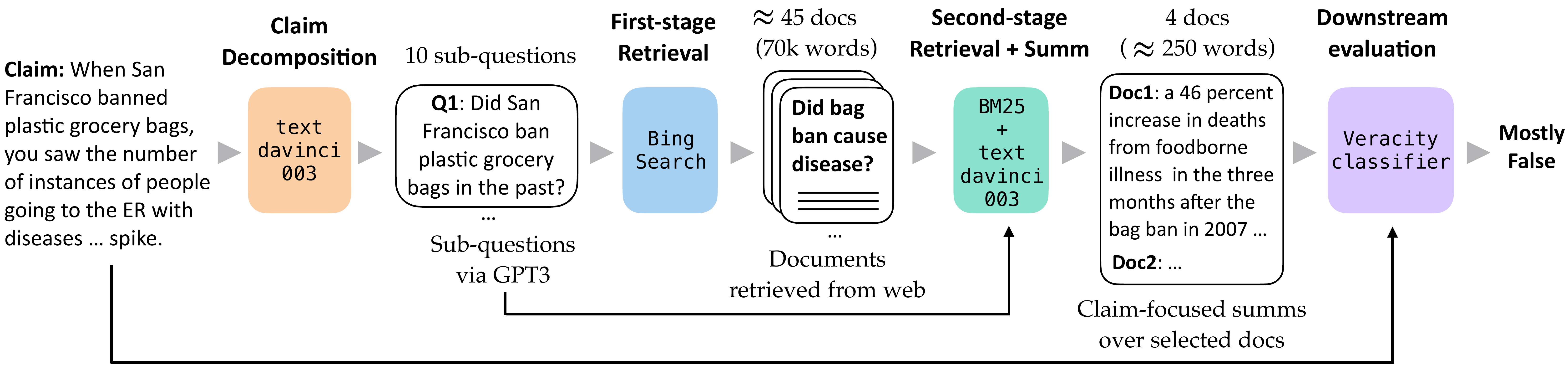}
  \caption{Overview of our pipeline: a claim is decomposed into yes/no subquestions (Sec.~\ref{sec:method-claim-decomp}), then we use the questions in two stages of retrieval (Sec.~\ref{sec:first-stage-retrieval} and Sec.~\ref{sec:method-second-stage-retrieval}) to select the most relevant paragraphs. Finally, we generate a claim-focused summary (Sec.~\ref{sec:method-claim-focused-summ}) and train a veracity classifier to get the veracity label (Sec.~\ref{sec:method-veracity-classification}). This filters contents irrelevant to the claim (see Appendix~\ref{appendix: information-compression} for details and an example in Figure~\ref{fig:case_study}).}\vspace{-1em}
  \label{fig:pipeline}
\end{figure*}
% Following the approach of~\citet{}, 
We present the first study of fact-checking political claims under a realistic retrieval setting. Our retrieval over the web is restricted to documents authored before the time of the claim and not sourced from fact-checking websites, as shown by the left side of Figure~\ref{fig:introduction}. We propose a pipeline (illustrated in Figure~\ref{fig:pipeline}) that builds upon prior work in fact checking as well as large language models~\cite{brown2020language} to handle the complexity of this setting. Our system first decomposes a claim into a series of subquestions~\cite{chen-etal-2022-generating,ousidhoum-etal-2022-varifocal}, targeting both explicit and implicit aspects of the claim. Each subquestion is fed into a commercial search engine to retrieve relevant documents, with the restrictions described above. Then, we conduct a second stage of fine-grained retrieval to isolate the most relevant portions of the documents. %Since many documents retrieved from search engines may be irrelevant to the claim or contain irrelevant information, 
Finally, we use state-of-the-art language models \cite{brown2020language,ouyang2022training} to generate claim-focused summaries from the retrieved content. These summaries can serve both as explanations for users as well as inputs to a classifier to determine the veracity based on these summaries.

%Our work addresses the realistic setting where evidence is not given to us (it has to be retrieved) \textbf{and} we are only allowed to consult raw evidence that would conceivably exist at a time that a fact was checked: articles that were This paper proposes a pipeline designed to retrieve raw evidence from the web for verifying complex political claims. 

Evaluating individual components of our pipeline is challenging due to the absence of gold annotations at each stage. We use automatic evaluation on the veracity classification performance, comparing to labels given by professional fact-checkers. We supplement this with a human study evaluating the claim-focused summaries for comprehensiveness and faithfulness. This evaluation counterbalances the subjectivity of the veracity judgments~\cite{lim2018checking} while shedding light on intermediate stages of the process.

We apply our pipeline to \textsc{ClaimDecomp}~\cite{chen-etal-2022-generating}, a dataset containing 1,200 real-world complex political claims with veracity labels. Performance on veracity classification shows that: (1) our retrieval setting is indeed much harder than ``unrestricted'' retrieval settings; (2) using web evidence leads to performance gains compared to automatic fact-checking without evidence; %, though there remains a significant gap between an oracle classifier performance based on human-written justifications
(3) the decomposition is crucial for obtaining high-quality raw documents from the web compared to using the original claim alone. Our human study further indicates that: (4) claim-focused summaries are mostly faithful and helpful for both machines and humans to fact-check a claim; (5) the retrieved evidence is often relevant to some aspects of the claim, but can rarely cover all aspects, suggesting that finding sufficient raw evidence in the wild is the core challenge in building automatic fact-checking systems. %We hope our work will spark NLP research in assisting fact-checkers in realistic scenarios.

\section{Background and Motivation}

% \paragraph{Fact verification} 
%Many of the widely used fact verification benchmarks, such as FEVER~\cite{thorne-etal-2018-fever}, HoVer~\cite{jiang-etal-2020-hover} and \textsc{VitaminC}~\cite{schuster-etal-2021-get}, focus on crowd-authored claims derived from Wikipedia. 
% For example, claims in the FEVER dataset typically require checking a single aspect like ``\emph{Oliver Reed was a film actor.}'' 
%These claims can be annotated at scale, but they do not reflect the complexities of real-world political claims.

Early NLP research on fact-checking political claims~\cite{vlachos-riedel-2014-fact, wang-2017-liar, rashkin-etal-2017-truth, volkova-etal-2017-separating, perez-rosas-etal-2018-automatic, dungs-etal-2018-rumour} typically considered using the claim alone as an input to an automated system. By not seeking evidence, systems judge the veracity of a claim mostly based on surface-level linguistic patterns rather than based on factual errors. Research that incorporates evidence either assumes access to justifications provided by fact-checkers~\cite{vlachos-riedel-2014-fact, alhindi-etal-2018-evidence, hanselowski-etal-2019-richly, atanasova-etal-2020-generating-fact} or evidence from \emph{unconstrained} retrieval~\cite{popat2017truth, popat-etal-2018-declare, augenstein-etal-2019-multifc}, which frequently yields evidence sets containing pages from fact-checking websites \citep{glockner-etal-2022-missing}. This does not reflect the difficulties in real-world evidence retrieval. \citet{fan-etal-2020-generating} explore generating questions to retrieve evidence from the web, but only evaluate their system with humans in the loop, who can aggressively filter irrelevant retrieval results. Contemporaneous to this work, \citet{schlichtkrull2024averitec} construct a dataset, AVeriTeC, using real-world claims and evidence retrieved from the web. Our method uses binary subquestions designed to target all needed aspects of factuality for a claim, whereas their questions are wh-questions optimized around retrieval, similar to QABriefs~\citep{fan-etal-2020-generating}.

% They also have human in the loop to ensure the retrieved evidence is sufficient to check the claim. However, in the real-world scenario, for many claims, no evidence can be found on the entire web.

To our knowledge, we present the first automatic fact-checking system with a realistic retrieval pipeline using evidence available at the time a claim was made. This presents a very challenging setting where many claims are not checkable. We therefore emphasize the evidence our system returns as a way of assisting human fact-checkers; we believe this realistic task setting and corresponding evaluation should be reused in future work.

%paves the way for work in more realistic settings and also
Our work shifts the focus away from the evaluation on classification accuracy alone. Accuracy on truth labels assigned by fact-checkers is a proxy metric we use to evaluate our systems. However, fact-checking experts argue that the task is too subjective and complex to be automated in the near term \cite{Graves2018,Nakov2021AutomatedFF}. Part of this arises from the fact that information needed to check claims is not always available on the web \cite{singh2021case}.
%Closed-book systems like GPT-4 cannot access this information at all, and even systems like WebGPT \cite{nakano2022webgpt} struggle to access all the requisite information; such systems are not yet optimized for issuing a large number of queries to address subtle, potentially subjective questions.
Our approach of returning information on a best-effort basis and providing evidence to enable humans to assist in the judgment can help overcome issues with returning judgments from error-prone AI systems \cite{bansaletal2021wholeparts, brand2022neural}.

%What is missing in the literature is the real in-the-wild evidence retrieval for automated fact-checking. In this work, we present the first fully automatic pipeline for verifying real-world claims by conducting constrained evidence retrieval from the web which closely resembles the real-world scenarios.

% \paragraph{Retrieval-augmented methods in NLP} 
% A wide range of NLP tasks benefits from knowledge augmentation from retrieval. Most prior work~\cite{} on augmenting retrieval studied question answering task. 
% \greg{discuss retrieval methods for QA I guess? say we are the first to do this for claim verification?}We only show three questions for simplicity.

\begin{figure}
  \centering
  \includegraphics[width=\columnwidth]{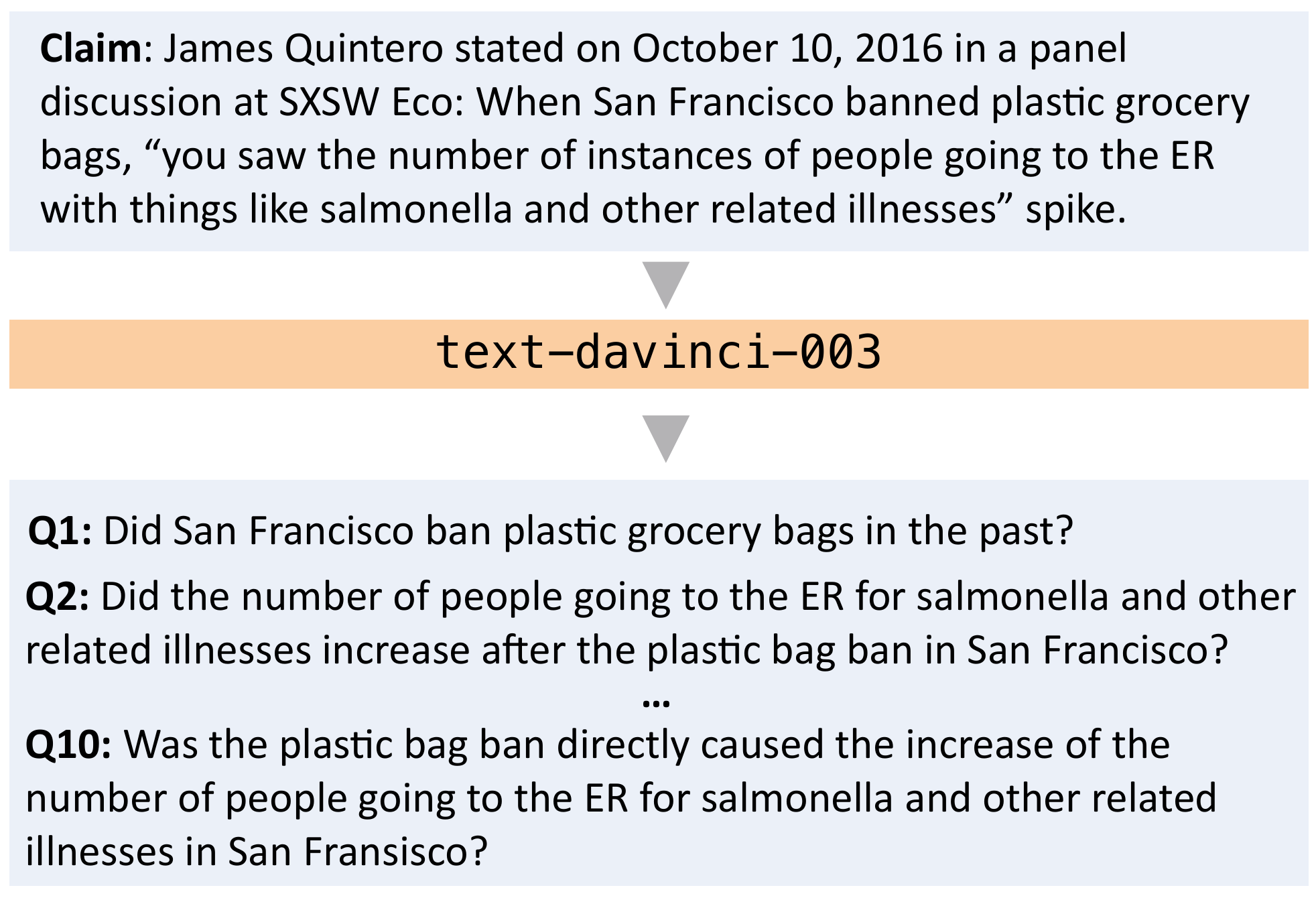}
  \caption{An example of our claim decomposition process: each claim is decomposed into ten subquestions. }
  \label{fig:claim-decomposition}
\end{figure}

\section{Methodology}

Our pipeline, shown in Figure~\ref{fig:claim-decomposition}, consists of five parts: claim decomposition, raw document retrieval, fine-grained retrieval, claim-focused summarization, and veracity classification. We describe each part below.% in this section. 

\subsection{Claim Decomposition} \label{sec:method-claim-decomp}
Given a real-world complex claim, we first decompose it into a set of yes/no questions for which the answers are useful to fact-check the claim. \citet{chen-etal-2022-generating, ousidhoum-etal-2022-varifocal} show that such decompositions are both helpful to retrieve relevant evidence and make veracity judgments.

For decomposition, we prompt a large-scale language model, \texttt{text-davinci-003}, with in-context examples.\footnote{During a pilot study, we compared the questions generated from \texttt{text-davinci-003} and the questions generated using the fine-tuned T5-3b model from~\citet{chen-etal-2022-generating} and we found that the questions generated by \texttt{text-davinci-003} are more diverse and comprehensive.} We carefully choose four input-decomposition pairs from the human annotations of~\citet{chen-etal-2022-generating} to form a few-shot prompt. We generate a set of questions through multiple rounds of sampling until we gather 10 different questions. An example decomposition is shown in Figure~\ref{fig:claim-decomposition}. For the full prompt, see Appendix~\ref{appendix:qg-prompt}.

\subsection{First-stage Retrieval}\label{sec:first-stage-retrieval}

For each question generated in the previous step, we feed it to a commercial search engine API to collect the relevant documents. 

\begin{figure}
  \centering
  \includegraphics[width=0.9\columnwidth]{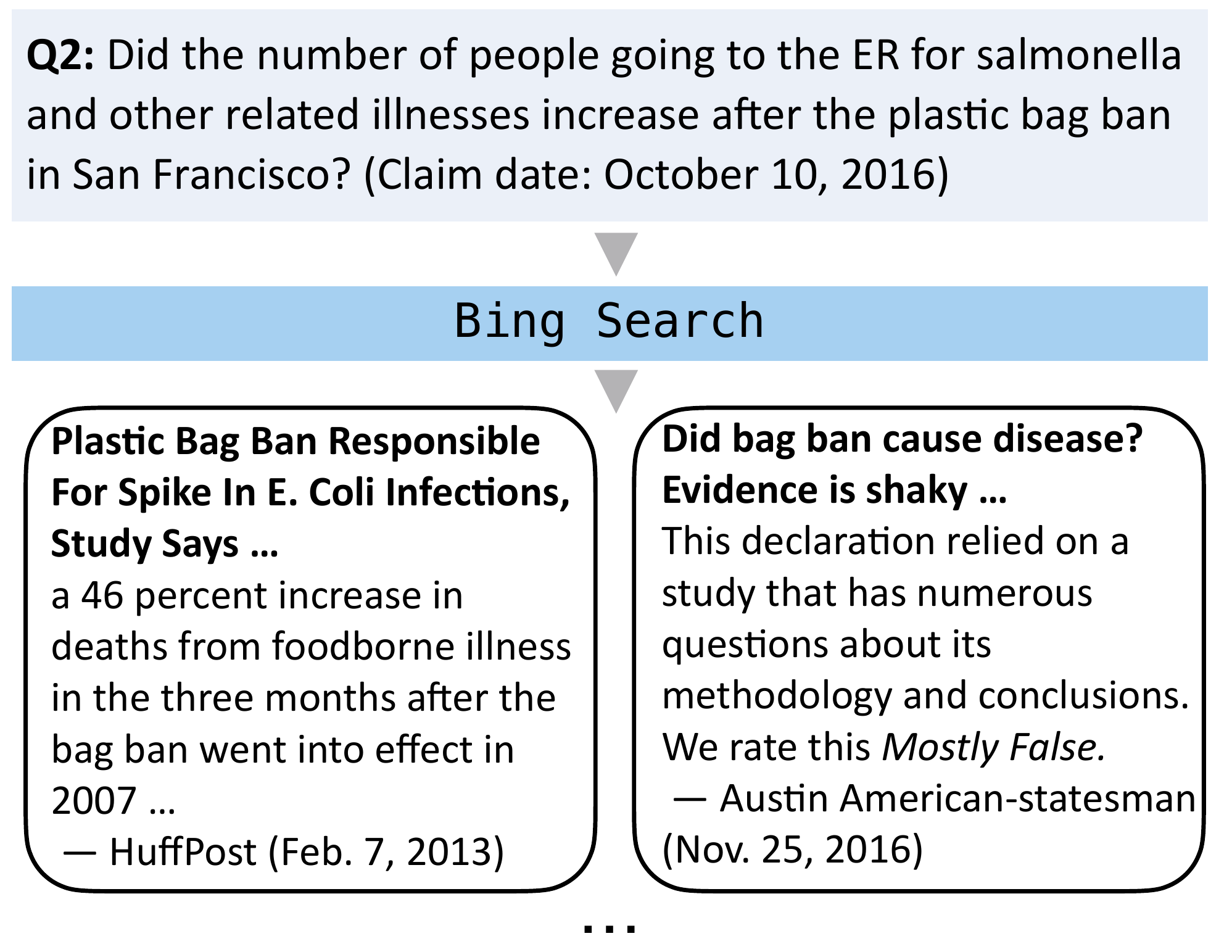}
  \caption{Two documents returned by searching Q2 (generated in step 1). The right page post-dates the claim by one month and directly cites a PolitiFact article, making it problematic to use as raw evidence.}
  \label{fig:doc-retrieval-via-timestamp}
\end{figure}

\paragraph{Temporal and Site Constraints} We assume that \textbf{a system should not be able to access pages published after the claim was made.} This condition matches real-time fact-checking scenario during a political speech. We place a \textbf{temporal constraint} on the system to reflect this. Next, to investigate how the presence of fact-checking websites affects the veracity judgment of a claim, we also place a \textbf{site constraint} to filter out the documents from fact-checking websites. Our list of fact-checking websites can be found in Appendix~\ref{appendix:websites-filtered}.  An example of the retrieved documents is shown in Figure~\ref{fig:doc-retrieval-via-timestamp}.

\begin{figure*}[t!]
  \centering
  \includegraphics[width=\textwidth]{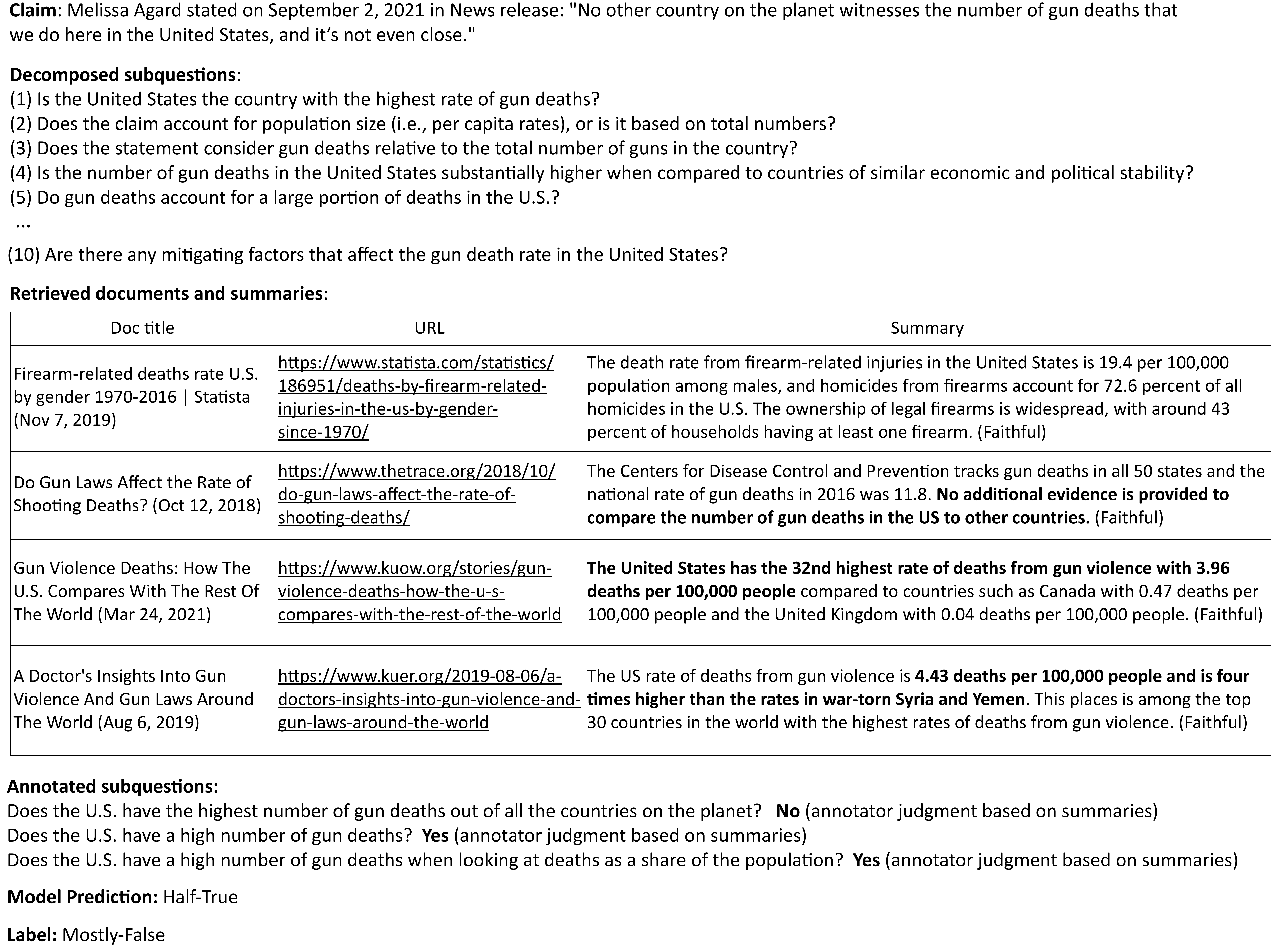}
  \caption{System outputs for an example picked from the dev set of \textsc{ClaimDecomp}: the claim is first decomposed into a set of yes/no questions and then the top four retrieved documents (through first and second stage retrieval) are summarized. Finally, a trained DeBERTa model makes a prediction regarding the four summarized documents. }
  \label{fig:case_study}
\end{figure*}

We use the Bing Search API,\footnote{\url{http://www.microsoft.com/en-us/bing/apis/bing-web-search-api}} and retrieve 10 documents per subquestion after filtering by the constraints. We extract the actual content from the page URLs using two tools: \texttt{html2text}\footnote{\url{https://github.com/Alir3z4/html2text/}} and \texttt{readability-lxml}.\footnote{\url{https://github.com/buriy/python-readability}} Approximately one-third of the URLs are protected\footnote{Paywall, PDFs, and anti-scraping measures.} and cannot be scraped.

Table~\ref{tab:web-retrieval-stats} contains the raw counts from web retrieval with and without the timestamp of a claim. These results underscore the importance of temporal filtering: we find little overlap between the two document sets by comparing the Jaccard distance between two sets of the retrieved URLs.

One challenge for the reproducibility of our work is that commercial search engines may return different results over time. In Section~\ref{sec:retrieval-instability}, we experiment with the same query set at different times. We find that the search results change over time: only 30\% of search output URLs overlap when queried two months apart. However, the veracity judgment classification result is not impacted much. %the stability of the Bing Search API as it is queried over time.

% In the ideal scenario, an automatic fact-checking system should be able to verify a claim as soon as it has been made. However, sometimes there is not enough evidence on the web right after the statement was made, and relevant contents arise afterward. We use search results in three different settings: \textbf{unconstrained}, \textbf{time-constrained (strict)} (only search documents that were available before the statement), \textbf{time-constrained (relaxed)} (search documents that were available one month after the statement).\gd{we need to update this para, right? doesn't match the results table} Therefore, we conduct two separate rounds of web searches with and without the timestamp of a claim.\footnote{We set the timestamp to be one month after the claim is made.}
%We combine the two sets of documents as the corpus for our experiments.

\begin{table}[t]
\small
\centering
\small
%\resizebox{0.5\textwidth}{!}{\begin{tabular}{ l  c c c }
\begin{tabular}{ l  c c c }
\toprule
	& \# retrieved  & \# scraped & \# words \\
 \midrule
w/ timestamp & 66.7 & 45.0 & 1,561 \\
w/o timestamp & 70.4 & 47.8 & 1,660 \\	
Jaccard score & 0.12 & 0.12 & - \\
\bottomrule
\end{tabular}
\caption{The statistics for the retrieved documents obtained through the first-stage retrieval after filtering the documents from fact-checking websites. Jaccard between these two sets show that incorporating the timestamp in retrieval makes a substantial difference.} 
\label{tab:web-retrieval-stats}
\end{table}

\subsection{Second-stage Retrieval} \label{sec:method-second-stage-retrieval}
Most of the documents collected from the previous step contain at most a few snippets relevant to the claim. However, as can be seen from Figure~\ref{fig:pipeline}, first-stage retrieval can easily result in tens of thousands of words of retrieved documents, which are costly to process with an LLM. Furthermore, even with state-of-the-art language models, it is hard to do complex reasoning over such long context~\cite{liu2024lost, levy2024same}. Thus, we conduct a second-stage retrieval to pick the most relevant text spans to the claim from the retrieved documents. Specifically, we segment the documents into text spans containing $k_1$ words with a stride of $\frac{1}{2}k_1$ words. Following~\citet{chen-etal-2022-generating}, we employ \texttt{BM-25} to retrieve the top-$K_1$ highest-scored text spans, expanding these spans with a $\pm$ $k_2$-word context. If two text spans overlap, they are merged to form a larger span. This process yields a set of ``documents'' ranked by the highest-scored text spans, of which we pick the top-$K_2$.

\subsection{Claim-Focused Summarization} \label{sec:method-claim-focused-summ}
Since the documents retrieved in the previous step can contain up to several thousand words, it becomes cumbersome for both humans and models to make a judgment based on them~\cite{stammbach2020fever}. Consequently, we prompt a large language model, specifically \texttt{text-davinci-003}, to summarize each retrieved document \emph{separately} with respect to the claim.\footnote{During a pilot study, we explored prompting \texttt{text-davinci-003} to generate one summary using all documents. However, it frequently went beyond simple summarization and produced verdicts such as ``therefore, the claim is refuted by the document,'' which were unreliable compared to using our veracity classifier.} Such single-document summarization has been shown to be robust on news articles \cite{goyal2022news,zhang2023benchmarking}.

%\paragraph{Producing summaries instead of judgments} %Few-shot vs. zero-shot summaries}

We investigate two types of prompts. For a \textbf{zero-shot} prompt, we instruct the model not to make any judgments about the stance of the given document. For a \textbf{few-shot} prompt, we select four documents and carefully write desired summaries. For documents that are not relevant to the claim, we write ``the document is not relevant to checking the claim'' as its desired output. We conduct human evaluation of the summary quality of different prompts in Section~\ref{sec:human-study-faithfulness}, where we find that few-shot prompting works better. See Appendix~\ref{appendix:summarization-prompt} for full prompts.

%We also compare against a variant of our system using a single summary, rather than one per document. We use zero-shot prompting for this \textbf{single-zero} variant.

\subsection{Veracity Classification}\label{sec:method-veracity-classification}
The final stage of our pipeline involves making a judgment based on the summaries generated in the previous stage. Unlike previous stages which use off-the-shelf tools, here we \textit{train} a DeBERTa-large~\cite{hedeberta} model\footnote{We also experimented with using ChatGPT as the veracity classifier. We describe results and analysis in Appendix~\ref{appendix:GPT-classifier}; we found it yielded worse performance than the fine-tuned model.} to perform a six-way veracity classification (true, mostly true, half true, barely true, false, and pants-on-fire).

\paragraph{Training} We run our pipeline over the training, development, and test data of \textsc{ClaimDecomp} and train on pairs of the form (claim+summary, label). Since the dataset is small, we train the classifier five times with different random seeds and report the test set performance using the model that achieves the best performance on the development set. 

% The input to the classifier is a concatenation of the claim and the summaries of the retrieved documents, while the output is one of the six labels. We use a classification head on the CLS token and train it with cross-entropy loss.

%Since we use few-shot prompting of large language models for both the decomposition and summarization phases, the only part of our system that requires training is the veracity classifier.
% We can train our model using the labels from \textsc{ClaimDecomp}; however, the inputs to the model are claim-focused summaries which have to be derived from our pipeline. 

\subsection{Final Pipeline}

Our complete pipeline's results when executed on an example are shown in Figure~\ref{fig:case_study}. We note that the question decomposition phase yields an overcomplete set of questions, including redundant ones. However, the final retrieved and summarized documents are able to shed light on the claim from several complementary perspectives. While the final veracity judgment does not exactly match the judgment from PolitiFact, reading the documents still gives an informed picture of the situation.

\section{Experimental Setup}
Our main automatic evaluation is on claim veracity prediction~\cite{wang-2017-liar}, evaluating our entire pipeline end-to-end. We will describe the human evaluation setup in Section~\ref{sec:human_eval}.
% end-to-end on this task.

\paragraph{Data} We use the data from \textsc{ClaimDecomp}~\cite{chen-etal-2022-generating} which contains 1,200 complex claims from PolitiFact (train: 800, dev: 200, test: 200). Each claim is labeled with one of the six veracity labels, a justification paragraph written by expert fact-checkers, and subquestions annotated by prior work.

% In section~\ref{sec:pipeline-ablation}, we investigate the difference between using our generated subquestions and that metadata.

\begin{table*}
\renewcommand{\tabcolsep}{1.4mm}
\footnotesize
\centering
%\resizebox{0.5\textwidth}{!}{\begin{tabular}{ l l c c c c }
\begin{tabular}{ c c  c c c c c c c c}
\toprule
      \multicolumn{2}{c}{Retrieval Constraint}   &         \multicolumn{4}{c}{Dev (N=200)} & \multicolumn{4}{c}{Test (N=200)} \\
Temporal & Site  & Acc & Soft Acc & Macro-F1 & MAE & Acc & Soft Acc & Macro-F1 & MAE \\
 \midrule
- & - & 50.5 & 88.5 & 47.5 & 0.62 & $\text{49.0}^+$ & $\text{86.0}^+$ & $\text{48.5}^+$ & $\text{0.68}^+$ \\
- & Non-FC & 37.5 & 76.5 & 38.6 & 0.94 & $\text{33.5}^+$ & $\text{75.0}^+$ & $\text{33.9}^+$ & $\text{0.95}^+$ \\
% Before & - & 42.0 & 83.0 & 42.7 & 0.80 & 39.5 & 79.0 & 39.8 & 0.86 \\
Before & - & 42.5 & 75.0 & 41.7 & 0.87 & $\text{33.5}^+$ & 72.0\phantom{$^+$} & $\text{38.0}^+$ & $\text{0.98}^+$ \\
Before & Non-FC & 40.5 & 76.5 & 41.4 & 0.87 & $\text{33.0}^+$ & $\text{74.5}^+$ & $\text{34.5}^+$ & $\text{0.99}^+$ \\
\midrule
\midrule
\multicolumn{2}{c}{Claim only} & 37.0 & 71.0 & 34.6 & 0.98 & 25.5 & 68.0 & 27.5 & 1.12  \\
\multicolumn{2}{c}{\textcolor{red}{Claim + Justification (oracle)}} & 52.5 & 88.5 & 54.5 & 0.64 & 57.5  & 93.0 & 57.8 & 0.50 \\

\bottomrule
\end{tabular}

\caption{Veracity classification performance with different retrieval constraints. The top block is our full system (\circled{B} setting in Table~\ref{tab:pipeline-ablations}) with constraints over what is retrieved. Red indicates using oracle information. ``$+$'' denotes that the results are statistically significant improvements ($p < 0.05$) compared to the results of Claim only on the test set.}

\label{tab:main_table}
\end{table*}
\paragraph{Hyperparameters} For the second-stage retrieval, we set $\text{top-}K_1=10$ (highest-scored text spans), $\text{top-}K_2=4$ (highest-scored documents), $k_1 = 30$ (chunk size), and $k_2 = 150$ (expansion parameter). See appendix~\ref{appendix:hyperparameters} for all hyperparameters.

\paragraph{Evaluation Metric} \label{sec:evaluation-metric}

We report accuracy (Acc), mean absolute error (MAE, on our 6-point scale), and Macro-F1. We also introduce soft accuracy (soft Acc), which is calculated by counting off-by-one errors on the six-point veracity scale (e.g., \emph{half true} instead of \emph{mostly true}) as correct, as veracity judgments are subjective.

\paragraph{Comparison Systems} For our \textbf{Claim-only} system, we concatenate the metadata, including the speaker and the venue of the claim, with the claim itself, and feed the resulting text into the classifier~\citep{wang-2017-liar}. This approach serves as a lower bound for the veracity classification. 

We extend the Claim-only baseline to \textbf{Claim+Justification} by appending the human-written justification paragraph, excluding the sentence containing the label, to the claim. This is an {oracle setting} to establish an upper bound for veracity classification.

%\paragraph{Retrieval constraints} By default, we use evidence retrieved under both sites and temporal constraints as described in Section~\ref{sec:first-stage-retrieval}. We experiment with four distinct variants by combining the two constraints (Section~\ref{sec:search_constraints}).

%\paragraph{Stage Ablation} We ablate the first-stage retrieval, second-stage retrieval, the summarization model, and the classifier to understand how each individual component contributes to the final performance (Section~\ref{sec:pipeline-ablation}).

\section{Automatic Evaluation: Claim Veracity}
\label{sec:eval_claim_veracity}

\subsection{Constrained vs. Unconstrained Search}
\label{sec:search_constraints}

We first situate our work with respect to baselines and past systems by varying the retrieval condition. We experiment with a \textbf{temporal} constraint, where pages must originate before the date of the claim, and a \textbf{site} constraint, where sites must be non-fact-checking (non-FC) sites. Even in the unconstrained setting, we exclude pages from PolitiFact (our dataset's source) to prevent label leakage.

The unconstrained setting corresponds to that used in MultiFC \cite{augenstein-etal-2019-multifc}. \textbf{MultiFC includes numerous documents that are filtered out by our constrained settings.} For each claim, they extract the top 10 pages from the Google search API. We find that 12,721 out of 15,379 claims (82.7\%) contain at least one page from our excluded website list and 24.4\% of the retrieved web pages are from fact-checking websites.

Table~\ref{tab:main_table} reports the performance of our system with various retrieval constraints. Comparing the performance of \emph{claim-only} and other models that use retrieval, we see a statistically significant\footnote{Throughout our study, we use paired bootstrap tests for statistical significance between the results.} improvement over all four of our metrics in nearly all settings, showing that \textbf{retrieving and summarizing evidence is helpful to predict the veracity label, even with constraints}. %We investigate more deeply which parts of our pipeline are responsible for this performance in the next section.

Second, we see \textbf{adding either temporal or site constraints dramatically reduces the performance}. This implies that retrieval over the web works largely because it retrieves fact-checks that were published after the claim was released, with synthesized evidence. We believe that future work on retrieval should use a constrained setting.

\begin{table*}
\renewcommand{\tabcolsep}{1.4mm}
\footnotesize
\centering
\begin{tabular}{l c c c   c c c c }
\toprule
 &\multicolumn{3}{c}{Evidence Generation}   & \multicolumn{4}{c}{Performance} \\
 & {FSR} & SSR & Summary  & Acc & Soft Acc & Macro-F1 & MAE \\
\midrule
%\multicolumn{8}{c}{Our system} \\
%\midrule

 \multicolumn{4}{c}{Claim only} & $\text{25.5}^+$ & $\text{68.0}^+$ & $\text{27.5}^+$ & $\text{1.12}^+$ \\
\multicolumn{4}{c}{\textcolor{red}{Claim + Justification}} &  $\text{57.5}^+$  & $\text{93.0}^+$ & $\text{57.8}^+$ & $\text{0.50}^+$ \\
\midrule
\multicolumn{8}{c}{Our Default System} \\ 

\midrule
\circled{B} & subQs  & subQs &  \texttt{zero-shot-003} & 33.0 & 74.5 & 34.5 & 0.99 \\

  \midrule
 \multicolumn{8}{c}{Ablation on first-stage retrieval} \\
 \midrule
\circled{1} & Claim &   &  & $\text{24.5}^+$ & 71.5 & $\text{18.0}^+$ & $\text{1.15}^+$ \\
\circled{2} &  Gold subQs  &  & & 27.5\phantom{$^+$} & 72.0 & $\text{28.1}^+$ & $\text{1.05}^+$ \\
% AnnotQs & GnrtQs & multi-zero & DeBERTa & Claim + Summ & & & & \\
\midrule
\multicolumn{8}{c}{Ablation on second-stage retrieval} \\
\midrule
\circled{3} & & Claim &  & 31.5 & 75.0 & 35.6 & 0.97 \\
\circled{4} && Gold subQs & & 31.5 & 73.0 & 35.4 & 1.03 \\
\textcolor{red}{\circled{5}} & &  \textcolor{red}{Justification} &   & 33.0 & 71.5 & 37.2 & 1.01 \\
\midrule
\multicolumn{8}{c}{Ablation on summarization} \\
\midrule
\circled{6}  & &  &  \texttt{few-shot-003}& 35.0 & 76.5\phantom{$^+$} & 36.2\phantom{$^+$} & 0.94\phantom{$^+$} \\
% \circled{6}  & &  &single-zero & 42.0 & 53.0 & 40.9 & 0.85 \\
\circled{7}  & &  & no summary (raw doc) & 29.0 & $\text{66.0}^+$ & $\text{26.3}^+$ &  {$\text{1.18}^+$}\\
% \midrule
% \multicolumn{9}{c}{Ablation on Decision Model} \\
% \midrule
% subQs & subQs & multi-zero & \texttt{Davinci-003} & Claim + Summ & & & & \\
% subQs & subQs & multi-zero & \texttt{ChatGPT} & Claim + Summ & & & & \\
\bottomrule
\end{tabular}
\vspace{-0.5em}
\caption{End-to-end fact-checking performance on the test set of \textsc{ClaimDecomp}. We ablate various stages of the model (FSR: first-stage retrieval; SSR: second-stage retrieval). Red indicates using oracle information. ``$+$'' denotes the result changes are statistically significant ($p < 0.05$) with respect to our default system.} 
\label{tab:pipeline-ablations}
\end{table*}

\subsection{Stage Ablations} \label{sec:pipeline-ablation}
We evaluate design choices in each stage of the pipeline to understand how each individual component contributes to the final performance. The results are shown in Table~\ref{tab:pipeline-ablations}. %Referring back to Figure~\ref{fig:pipeline}, we modify the following steps of the pipeline:
    % \item \textbf{First-stage retrieval:} Rather than retrieving with subquestions (\textbf{subQs}), we instead perform our search with the raw claim (\textbf{Claim}) and the gold subquestions annotated in ClaimDecomp.
    % \item \textbf{Second-stage retrieval:} 
    % \item \textbf{Summ}: 

%We have the following observations:

\paragraph{First-stage Retrieval: subquestions vs. original claim} Using the original claim instead of the generated subquestions as an input to web search (\circled{B} vs. \circled{1}) results in a notable decrease in performance. The subquestion set encompasses multiple aspects of the claim, enabling the search engine to locate relevant information more easily across separate search queries. Comparing \circled{B} and \circled{2}, we see using the gold subquestions actually yields worse performance than our predicted subquestions. This could be because we predict 10 subquestions, potentially garnering more relevant data than the 3 (on average) gold subquestions~\citep{chen-etal-2022-generating}.

\paragraph{Second-stage Retrieval} Rather than retrieving with subquestions (\textbf{subQs}), we instead perform our search with the raw \textbf{Claim} (\circled{3}), \textbf{Gold subQs} from \textsc{ClaimDecomp} (\circled{4}), or \textbf{Justification} (\textcolor{red}{\circled{5}}), which uses oracle information. Different queries yield only slight differences in performance and none of them is statistically significant, even when \textcolor{red}{\circled{5}} uses the human-written justification. We believe this is because we expand the retrieved text span by a context window ($\pm150$ words). As a result, this retrieval step does not need to be very precise to capture the relevant information.

\paragraph{Claim-focused Summarization} We compare \textbf{zero-shot} (\circled{B}) and \textbf{few-shot} (\circled{6}) prompts for generating the summary; \textbf{no summary} (\circled{7}) directly feeds the text spans from second-stage retrieval to the veracity classifier. System \circled{7} shows the worst performance across all metrics, suggesting that summarization matters. This may result from two primary factors: (1) The document length exceeds the context window capacity of DeBERTa, causing crucial information to be truncated. (2) our veracity classifier cannot easily discern the most relevant information given a large amount of context. Differences in the prompt (\circled{B} and \circled{6}) do not impact veracity classification results much but have differences under human inspection, which we discuss in the next section.% on our human evaluation in the next section.

\subsection{Stability of First-stage Retrieval}\label{sec:retrieval-instability}
As commercial search engines evolve over time, we conduct experiments to explore the reproducibility of our first-stage retrieval step. We use the default system setting in Table~\ref{tab:pipeline-ablations} and conducted three rounds of retrieval at $T=0$, $T=\textrm{1 week}$, and $T=\textrm{2 months}$. We evaluate the Jaccard similarity of the sets of URLs retrieved from our queries to understand how much changes in the Bing API and the broader web change our results. We also evaluate the veracity of our system. Note that this Jaccard similarity is between the members of the URL sets (i.e., the URLs themselves), not capturing any lexical or domain similarity of the URLs.

Results are shown in Table~\ref{tab:retrieval-stability}. A noticeable trend is a decline in the Jaccard score between varying retrieval rounds over time. However, this decrease does not significantly impact the models' efficacy in the veracity assessment. 

We caution that as the time gap increases, the set of documents retrieved from the Bing Search API could become considerably different, posing a challenge to consistently benchmark retrieval performance using commercial search engines. Therefore, we advocate for future research to focus on developing a comprehensive yet challenging document set that could be publicly released as a benchmark to spur research.

\begin{table}[t]
\small
\centering
\renewcommand{\tabcolsep}{1.4mm}
\begin{tabular}{ l c c c c c }
\toprule
 & Overlap  & Acc & Soft-Acc & Ma-F1 & MAE \\
 \midrule
 Ours  & -  & 33.0 & 74.5 & 34.5 & 0.99 \\
  1 week  & 0.48 & 33.5 & 74.0 & 36.8 &  0.98 \\
  2 months & 0.30 & 29.5 & 73.5 & 32.3 & 1.03 \\
\bottomrule
\end{tabular}
\caption{Model performance with respect to different rounds of retrieval at intervals of one week and two months. The overlap between ``Ours'' and subsequent document sets, measured with Jaccard score, decreases as the time gap increases. However, none of the changes in our downstream metrics is statistically significant.} 
\label{tab:retrieval-stability}
\end{table}

% We recruit annotators from Mechanical Turk following the same protocol of the faithfulness study. We select 15 workers and conduct a 3-way annotation over we picked for the human study of faithfulness. 

\section{Human Evaluation of Summaries}
\label{sec:human_eval}

Summarizing documents from web search with large language models improves the performance of our fact-checking pipeline. However, these models can generate untruthful content~\cite{bommasani2021opportunities, chowdhery2022palm, ouyang2022training}. Furthermore, as pointed out by~\citet{lim2018checking}, the accuracy of veracity classification alone does not entirely reflect the system's overall effectiveness, as certain labels such as ``false'' and ``barely-true'' may be ambiguous. We believe the true measure of our system's utility lies in the full package of summarized evidence it returns rather than just the accuracy of the veracity label. Therefore, we carry out two human studies, on comprehensiveness and faithfulness, to better understand intermediate outputs of the system.

\paragraph{Setting} We randomly pick 50 claims which contain 200 document-summary pairs from the development set of \textsc{ClaimDecomp} and run two human evaluation studies on this set. For each task, we recruited annotators from Amazon Mechanical Turk with a qualification test. In total, we recruited 17 worker for the faithfulness study and 15 workers for the comprehensiveness study. The details about crowdsourcing can be found in Appendix~\ref{appendix:human_study}. 

% that provides them five evaluation items, asking for both labels and justifications for their label. We select workers that get more than three out of five examples correct and provide reasonable explanations. We had a total of 17 workers for the faithfulness study and a total of 15 workers for the comprehensiveness study, and collect three annotations for each example. 

\paragraph{Comparison Systems} We compare the summaries generated from two prompts, \texttt{zero-shot-003} and \texttt{few-shot-003}, on GPT-3.5 (davinci-003). For the faithfulness study, we also compare the summaries generated through with zero-shot prompt on an earlier GPT model (davinci-001) (\texttt{zero-shot-001}) to see how the faithfulness varies for different models. 

\begin{table}[t]
\small
\centering
\resizebox{0.49\textwidth}{!}{\begin{tabular}{ l c c c c c }
\toprule
Summ-type & F & Minor & Major & NF & Avg score \\
 \midrule
 \texttt{zero-shot-001} & 65.8\% & 9.2\% & 20.0\% & 5.0\% & 3.45 \\
 \texttt{zero-shot-003} & 66.0\% & 18.0\% & 16.0\% & 0.0\% & 3.50 \\
 \texttt{few-shot-003} &  82.5\% & 6.5\% & 8.5\% & 2.5\% & 3.69 \\

% \texttt{zero-shot-001} & 79 & 24 & 11 & 6 &  \\
%  \texttt{zero-shot-003} & 132 & 36 & 32 & 0 & 3.50 \\
%  \texttt{few-shot-003} &  165 & 13 & 17 & 5 & 3.69 \\

\bottomrule
\end{tabular}}

\caption{Faithfulness Human Evaluation ($N=200$). ``F'' denotes that the summary is factual and ``NF'' denotes that the summary is completely wrong. Few-shot prompting helps the model make fewer factual errors.} \vspace{-0.6em}
\label{tab:human-study-faithfulness}
\end{table}

\subsection{Faithfulness Evaluation} \label{sec:human-study-faithfulness}
\paragraph{Goal} We assess the frequency and degree to which the language model generates untruthful content during query-focused summarization. For each document and summary pair, annotators choose one of four labels below (see appendix~\ref{appendix:unfaithful_summaries} for examples):
\begin{itemize}[nosep,leftmargin=*]
    \item \textbf{Faithful:} the summary accurately represents the meaning and details of the original document. 
    \item \textbf{Minor Factual Error:} some details are not aligned with the original document, but the overall message remains intact.
    \item \textbf{Major Factual Error:} there are factual errors that result in the summary misrepresenting the original document.
    \item \textbf{Completely Wrong:} the language model hallucinates content that completely alters the meaning of the original document. 
\end{itemize}
In addition to selecting a label, we ask annotators to provide a natural language justification for their choices. The annotations agree with a Fleiss Kappa score of 0.30. While this number is somewhat low, when we evaluated their justifications and we find many of the disagreements are because of subjectivity on the extent of factual error. We compute a consensus annotation via majority vote. We assign numerical scores to each label, where ``Faithful'', ``Minor'', ``Major'', and ``Completely Wrong'' correspond to 4, 3, 2, and 1 respectively and report average values. If all annotators disagree, we compute the average score and return the label that is nearest to the average score as a consensus. 
%Note that we do not ask the annotator to assess the quality of the generated summary.

%Human evaluation results on the same 200 document-summary pairs from 50 claims we randomly picked from zero-shot and few-shot summaries based on \texttt{text-davinci-003}
\paragraph{Results}

The results are shown in Table~\ref{tab:human-study-faithfulness}. We see that \textbf{few-shot prompting substantially decreases the chance of hallucinations in the summaries}. When combining ``Factual'' and ``Minor'', we see 89\% of the summaries are good enough to be used as evidence for the classifier. Additionally, by checking the unfaithful summaries, we find that they do not consist of useful hallucination like making a veracity judgment based on the parametric knowledge. Comparing the performance of \texttt{zero-shot-001} and \texttt{zero-shot-003}, we find that the weaker model makes more major factual errors. Together, they indicate that with stronger models and better prompts, we may expect these summarization models to improve further.

\subsection{Comprehensiveness Evaluation} \label{sec:human-study-comprehensiveness}
\paragraph{Goal} We aim to measure the extent to which the claim-focused summaries are able to address the claim. This is subjective and difficult task to evaluate. Here, we leverage the human-annotated yes/no subquestions from \textsc{ClaimDecomp} as a proxy for evaluating the comprehensiveness of our summaries: if provided summary can help humans to answer more of these yes/no questions, we deem the summary to be more comprehensive. 

In this task, annotators are given a summary / subquestion pair and label subquestion as ``answerable'', ``partially answerable'',\footnote{Sometimes the questions cannot be directly answered but can be inferred from the content of the summaries, or the summary at least contains relevant information. In such cases, we ask annotators to choose ``partially answerable''.} or ``unanswerable'', and additionally provide yes/no answer if the question is labeled as ``answerable''. Annotators were also asked to provide natural language justification for their answers. We collect this annotation on 161 questions associated with 50 claims. The annotations agree with a Fleiss Kappa score of 0.32. 
% for each subquestion, we ask annotators to determine whether the question is answerable or not. If the question is deemed answerable, we ask the annotator to indicate whether the answer is ``Yes'' or ``No'' based on their best judgment. 

\paragraph{Results}
The results are presented in Table~\ref{tab:human-study-comprehensiveness}. We see that zero-shot summaries yield more answerable questions than few-shot summaries. However, faithfulness evaluation hints that this is caused by hallucinations in zero-shot summaries; the system imputes information that seems to help, but which is not supported by the document.

Nevertheless, the few-shot summaries allow us to partially address over 60\% of the \textbf{gold annotated} subquestions derived from the PolitiFact justification. We find this result encouraging: even though the system does not have access to these (often subtle) factors, it can retrieve information to enable a human annotator to make a judgment about them.

\begin{table}[t]
\small
\centering
\begin{tabular}{ l c c c }
\toprule
Summ-type & Ans & Partially Ans & UnAns \\
 \midrule
 % first page & & & \\
 \texttt{zero-shot-003} & 47.8\% & 22.4\% & 29.8\%  \\
 \texttt{few-shot-003} &  42.9\% & 21.1\% & 36.0\%  \\

 % \texttt{zero-shot-003} & 77 & 36 & 48  \\
 % \texttt{few-shot-003} &  69 & 34 & 58  \\
\bottomrule
\end{tabular}
\caption{Human evaluation results on 161 subquestions from the same 50 claims we picked for the human study on faithfulness. ``Ans'', ``Partially Ans'', and ``UnAns'' denote the number of questions that are answerable, partially answerable, and unanswerable.} 
\label{tab:human-study-comprehensiveness}
\end{table}

\begin{table}
\small
\centering
\begin{tabular}{ l c c c  c  }
\toprule
 & Faithful & Minor & Unfaithful & Total \\
 \midrule
 Ans  & 4 & 2 & 0 & 6 \\
 Partially Ans  & 6 & 1 & 1 & 8 \\
 Partially UnAns & 13 & 5 & 11 & 30 \\
 UnAns & 5 & 1 & 0 & 6 \\
 \midrule
Total & 28 & 10 & 12 & 50 \\
\bottomrule
\end{tabular}

\caption{\textbf{Claim-level} statistics of \texttt{few-shot-003} taking both faithfulness and comprehensiveness into consideration. ``Unfaithful" label aggregates ``Major Error" and ``Completely Wrong" labels. The claim-level labels are derived from the sub-parts as defined in section~\ref{sec:human-study-combined}. } 
\label{tab:human-study-combined}
\end{table}

\subsection{Combined Evaluation} \label{sec:human-study-combined}
While in previous section we evaluated faithfulness and comprehensiveness separately, here we conduct a \textbf{claim-level} evaluation: how many claims can be comprehensively addressed with a set of faithful summaries? We label a claim as \textbf{answerable} if all of its subquestions are answerable. %If all subquestions are at least partially answerable, the claim is labeled as \textbf{partially answerable}. When only some subquestions are partially answerable, the claim is categorized as \textbf{partially unanswerable}. 
If all subquestions are unanswerable the claim is \textbf{unanswerable}. Otherwise, we label claim as \textbf{partially unanswerable}. For claim-level faithfulness, we apply the same principles: a claim is faithful is all summaries are faithful, otherwise it is either unfaithful or contains minor factual errors. Table~\ref{tab:human-study-combined} shows the results by combining the two factors. We see that addressing every aspect of complex claims is still challenging: 36 out of 50 claims contain at least one unanswerable question. For claims that can be fully addressed (all questions are either answerable or partially answerable), only 1 out of 14 contains a major factual error in the summary. %the retrieved documents.

\section{Related Work}
% \paragraph{Fact-checking with evidence} 
% Early\gd{copy-paste this into section 2} work focusing on checking real-world complex claims does not use supporting evidence beyond the claim itself~\citep{wang-2017-liar, rashkin-etal-2017-truth, volkova-etal-2017-separating, perez-rosas-etal-2018-automatic, dungs-etal-2018-rumour}. However, such systems fail to identify well-presented misinformation like machine-generated claim~\citep{schuster-etal-2020-limitations}. Research incorporating evidence often assumes having access to a set of human-written fact-checking articles~\citep{vlachos-riedel-2014-fact, alhindi-etal-2018-evidence, hanselowski-etal-2019-richly, atanasova-etal-2020-generating-fact, chen-etal-2022-generating} or evidence from \emph{unconstrained} retrieval~\cite{popat2017truth, popat-etal-2018-declare, augenstein-etal-2019-multifc}, which frequently yields evidence sets containing information from fact-checking websites themselves.

% To our knowledge, our work is the first to build a fully automated pipeline from retrieving evidence in the wild to assign the veracity and provide a set of summarized documents as justification. Our approach is most relevant to ~\citet{fan-etal-2020-generating} in which they also retrieve the evidence from web but they use human to do the retrieve and make judgment about the claim.

\paragraph{Retrieval augmented models}
Prior work has shown that a variety of NLP tasks could benefit from incorporating a retrieval component. Such tasks mainly include question answering~\citep{chen-etal-2017-reading, kwiatkowski-etal-2019-natural, karpukhin-etal-2020-dense, khattab-etal-2021-relevance,nakano2021webgpt}, text generation~\citep{lewis2020retrieval, shi2023replug, ram2023context}, language modeling~\citep{guu2020retrieval, Khandelwal2020Generalization, zhong-etal-2022-training}, and dialog~\citep{moghe-etal-2018-towards, fan-etal-2021-augmenting, thoppilan2022lamda}.

Most of these work assume having access to a fixed corpus, however, for the task of real-world fact-checking, no such corpus exists. In this work, we follow WebGPT~\citep{nakano2021webgpt} and use Bing Search API to retrieve evidence from the wild web. Recent LLM agents such as Bing Chat and Google Bard follow this paradigm, so we believe these directions will be relevant for future work.

\paragraph{Question decomposition} has been shown to be effective in evidence retrieval and question understanding for complex question answering~\citep{talmor-berant-2018-web, min-etal-2019-multi, qi-etal-2019-answering, perez-etal-2020-unsupervised, wolfson-etal-2020-break, geva-etal-2021-aristotle}. Question generation has also been shown to play a useful role in retrieval pipelines in open-domain QA \cite{sachan-etal-2022-improving}. In more recent research, it was demonstrated by ~\citet{chen-etal-2022-generating} that such decompositions can also aid in retrieving evidence to assess complex claims and make veracity judgment. This observation is consistent with concurrent studies on fact-checking text generation outputs~\citep{Gao2022RARRRA,chen2022propsegment,liu2022revisiting} and Wikipedia~\citep{kamoi2023wice}. %We show the value of question decomposition in retrieving articles with search engine API.%Our work resembles the similar idea by decomposing complex claims into a set of questions for better evidence retrieval.

\section{Conclusion}
We introduce a pipeline for realistic, automated fact-checking of complex political claims by retrieving raw evidence from web documents, improving final fact checking accuracy by integrating retrieved evidence. Our pipeline show promising results on the \textsc{ClaimDecomp} dataset. Yet, web search often cannot surface all the pieces of information necessary to verify a given claim. 
This work emphasizes the challenges of evidence retrieval in real-world scenarios and underscores the need for a human-in-the-loop fact-checking system.

\section*{Limitations and Future Directions}
\paragraph{Performance is bottlenecked by the first-stage retrieval.} The results in the last section show that 36.0\% of questions are unanswerable using our most faithful claim-focused summaries. By investigating the unanswerable cases, we see that the following cases lead to retrieval failure: (1) no relevant information is available on the web except the fact-checking websites. These claims can be onerous to check, such as requiring talking to or emailing specific people to check facts. Those cases are beyond the scope of this work and we think a system doing triage for the claims, would be promising for future work. (2) No relevant subquestions are generated or the subquestions are not well decontextualized~\cite{choi-etal-2021-decontextualization}. In such cases, a stronger question generation model or decontextualization model can help further. 

\paragraph{The need of human-in-the-loop fact-checking.} To address the failures in the first-stage retrieval and the potential errors in the summarization stage, we envision a human-in-the-loop fact-checking system. This system begins with the automated pipeline presented in this paper, which provides fact-checkers with summarized documents and judgments. If the fact-checkers deem these documents unsatisfactory, the system reveals the subquestions used for evidence retrieval, allowing fact-checkers to rerun the search. The system then retrieves additional documents and generates updated summaries. This iterative process continues until the fact-checkers are satisfied with the retrieved evidence. Moreover, the system could further learn from the fact-check feedback to improve itself: for example, the system could learn what questions are important to retrieve good evidence and what questions are not according to the fact-checker. In general, we believe such systems will be necessary, but developing them is outside of the scope of this work.

\paragraph{Scope of facts checked.} Our work only addresses English-language political claims. Misinformation in other languages is a crucial problem that we believe future work should address. Moreover, even within English, there is a strong need for fact-checking systems that can address other kinds of claims that have a different distribution; for example, claims from social media, which are often embedded in images or memes. Nevertheless, we believe the decomposition and retrieval approach here can play a role in such systems as well.

\section*{Acknowledgments}

This work was partially supported by NSF CAREER Award IIS-2145280, by Good Systems,\footnote{https://goodsystems.utexas.edu/} a UT Austin Grand Challenge to develop responsible AI technologies, and by grants from Salesforce Inc. and Open Philanthropy. We thank the UT Austin NLP community for feedback on the earlier drafts of the paper.

\bibliography{anthology,custom}

\appendix \label{sec:appendix}

\section*{Appendix}

\section{Experimental Details}
\subsection{List of Websites being Filtered}\label{appendix:websites-filtered}
\begin{itemize}
    \item \url{www.politifact.com}
    \item \url{www.snopes.com}
    \item \url{www.factcheck.org}
    \item \url{www.washingtonpost.com/news/fact-checker/}
    \item \url{www.apnews.com/hub/ap-fact-check}
    \item \url{www.fullfact.org}
    \item \url{www.reuters.com/fact-check}
\end{itemize}
We also filter the URLs that contain ``fact-check'' or ``factcheck''; we also filter any PDF files and videos.

\subsection{Question Generation Prompt and Deduplication}\label{appendix:qg-prompt}

% \paragraph{Question generation prompt}

\begin{figure}[t!]
  \centering
  \includegraphics[width=0.5\textwidth]{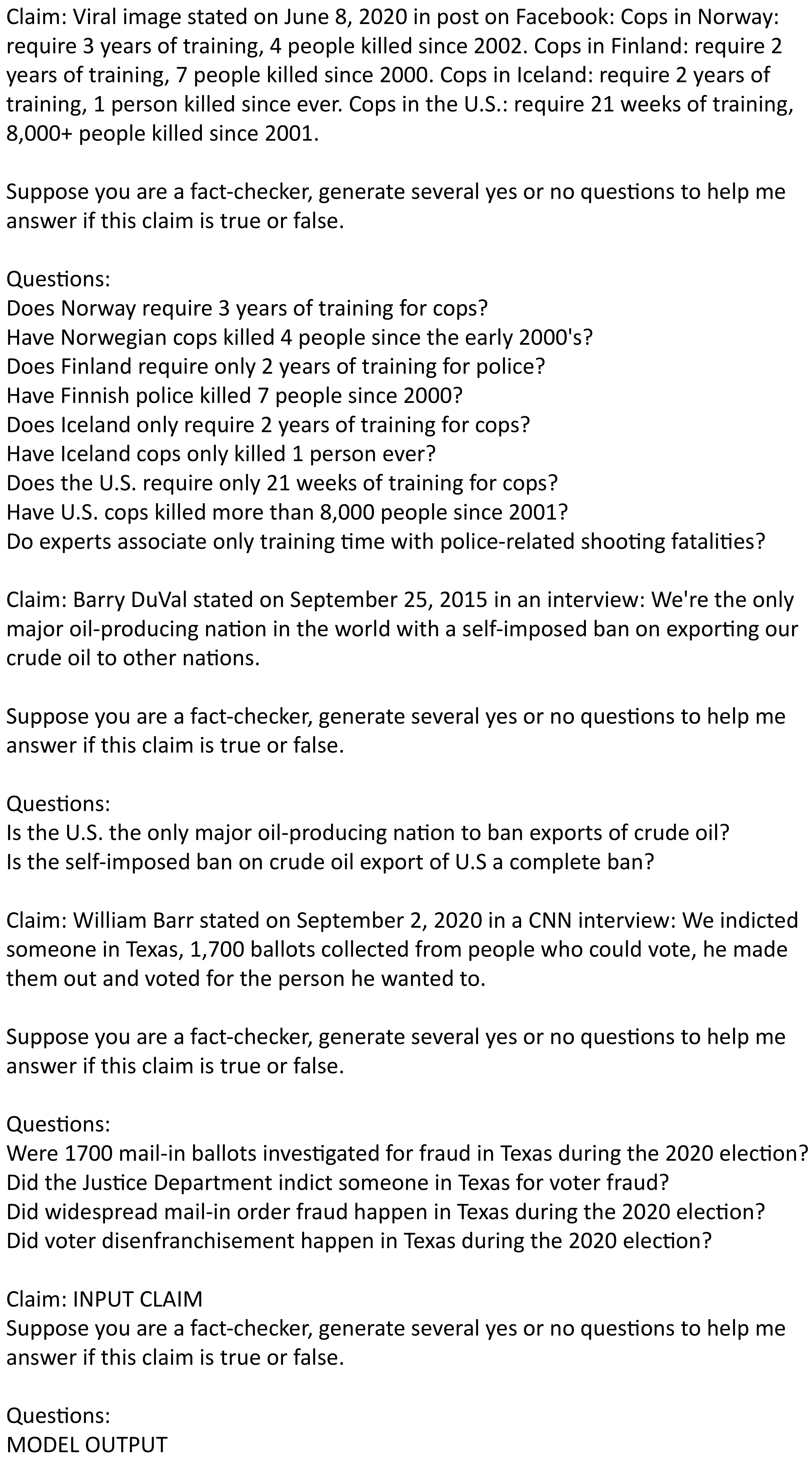}
  
  \caption{Few-shot prompt we used to generate subquestions in this paper.}
  \label{fig:qg-prompt}
\end{figure}

The prompt we used to generate the questions is shown in Figure~\ref{fig:qg-prompt}. Since the generated question set sometimes contains duplicates, we delete the duplicated questions according to the exact string match.

% \subsection{Document Retrieval Statistics with Temporal Constraints} \label{appendix:doc_stats_with_temporal_constraint}

% In practice, we implement the temporal constraint by adding a timestamp to the search engine. Table~\ref{tab:web-retrieval-stats} shows the statistics of the retrieved documents by two rounds of web retrieval, with and without the timestamp of a claim. We find little overlap between the two document sets by comparing the Jaccard distance between two sets of the retrieved URLs. Furthermore, approximately one-third of the URLs are protected\footnote{Paywall, PDFs, and anti-scraping measures.} and cannot be scraped as shown in Table~\ref{tab:web-retrieval-stats}. 

% \paragraph{Question deduplication}

% Since the generated question set sometimes contains duplicates, we delete the duplicated questions as follows: for each question pair, we calculate the similarity using ROUGE-1, DeBERTa-Large trained on Quora Question Pairs (QQP) dataset, and SimCSE. We take the average of the three values. If the average is greater than or equal to 0.5, we consider the two questions equal and take the maximum of the three values. If the average is less than 0.5, we consider the two questions not equal and take the minimum of the three values. With these selected values, we form a proximity matrix and perform agglomerative clustering. Once clustering is done, we select one question from each cluster.

\subsection{Question-focused Summarization Prompt}\label{appendix:summarization-prompt}

\begin{figure}[t!]
  \centering
  \includegraphics[width=0.5\textwidth]{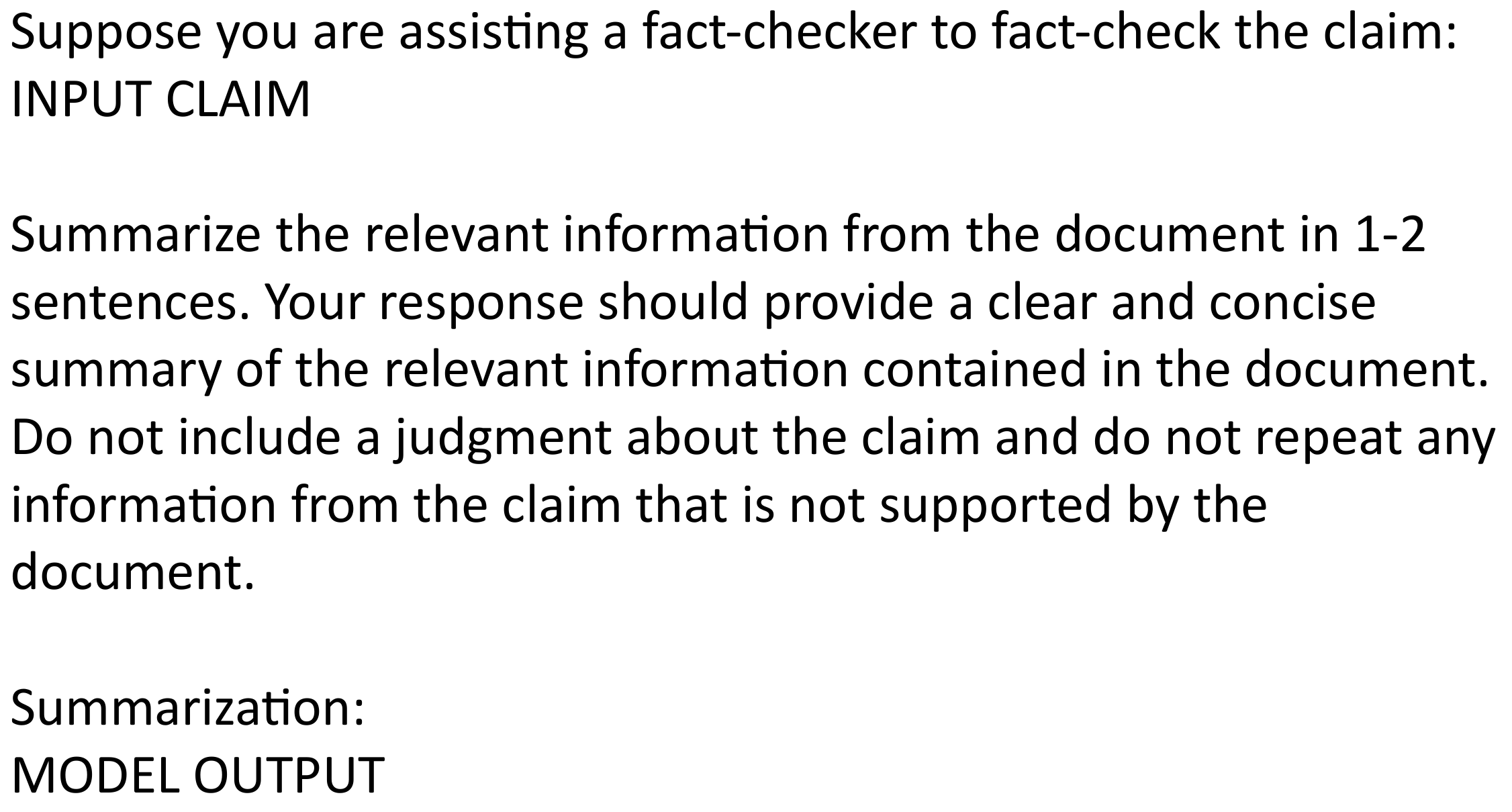}
  
  \caption{Zero-shot prompt we used to generate the claim-focused summaries in this paper.}
  \label{fig:doc-summary-zero-shot}
\end{figure}

\begin{figure*}[t!]
  \centering
  \includegraphics[width=\textwidth]{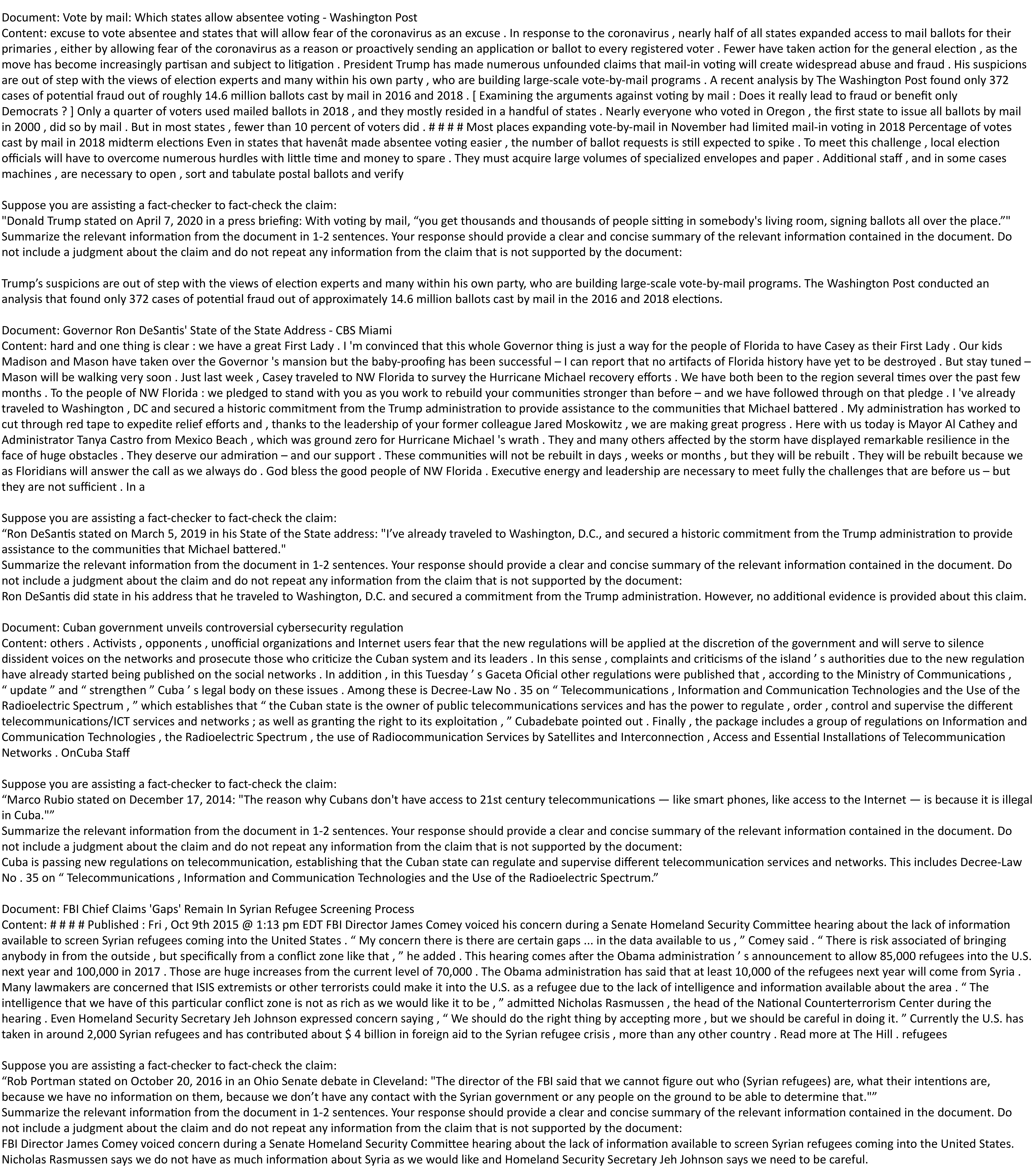}
  
  \caption{Few-shot prompt we used to generate the claim-focused summaries in this paper.}
  \label{fig:doc-summary-few-shot}
\end{figure*}

The zero-shot and few-shot prompts we used to generate the claim-focused summaries are shown in Figure~\ref{fig:doc-summary-zero-shot} and Figure~\ref{fig:doc-summary-few-shot} respectively.

\subsection{Hyperparameters of Veracity Classifier} \label{appendix:hyperparameters}

\begin{itemize}
    \item Model: DeBERTa-large
    \item Batch size: 32
    \item Max sequence length: 512
    \item Epochs: 25
    \item Initial learning rate: 3e-5
    \item Optimizer: Adam with linear decay
    \item Metric for selecting best dev model: MAE
    \item Random seed of 5 runs: 290032, 33432, 7876, 366, 77
    \item Training device: NVIDIA-A6000
\end{itemize}

% \section{Reproducibility of First-stage Retrieval} \label{appendix:search_stability}
% We conduct experiments to explore the stability and reproducibility of our first-stage retrieval step. We conducted three rounds of retrieval at $T=0$, $T=\textrm{1 week}$, and $T=\textrm{2 months}$. We evaluate the Jaccard similarity of the sets of URLs retrieved from our queries to understand how much changes in the Bing API and the broader web change our results. We also evaluate the veracity of our system. Note that this Jaccard similarity is between the members of the URL sets (i.e., the URLs themselves), not capturing any lexical or domain similarity of the URLs.

% Results are shown in Table~\ref{tab:retrieval-stability}. A noticeable trend is a decline in the Jaccard score between varying retrieval rounds over time. However, this decrease does not significantly impact the models' efficacy in the veracity assessment. 

% We caution that as the time gap increases, the set of documents retrieved from the Bing Search API could become considerably different, posing a challenge to consistently benchmark retrieval performance using commercial search engines. Therefore, we advocate for future research to focus on developing a comprehensive yet challenging document set that could be publicly released as a benchmark to spur research.

\section{Information Compression through the Pipeline} \label{appendix: information-compression}

Our pipeline progressively refines the crucial data needed to validate a claim. Table~\ref{tab:information-compression} demonstrates the average count of unique documents and the total word count in these documents after each phase of our pipeline under both temporal and site constraints.

\begin{table}[t]
\small
\centering
\begin{tabular}{ l c c c  }
\toprule
 & First-stage  & Second-stage & Summ \\
 \midrule
\# documents  & 45.0  & 7.7 &  4.0 \\
\# words  & 70,245 & 2,710 &  251 \\
\bottomrule
\end{tabular}
\caption{Average number of unique documents and average number of words in total from those documents after each stage of our pipeline.} 
\label{tab:information-compression}
\end{table}

\section{Human Study} \label{appendix:human_study}

\subsection{Examples of Unfaithful Summaries} \label{appendix:unfaithful_summaries}

\begin{figure*}
  \centering
  \includegraphics[width=\textwidth]{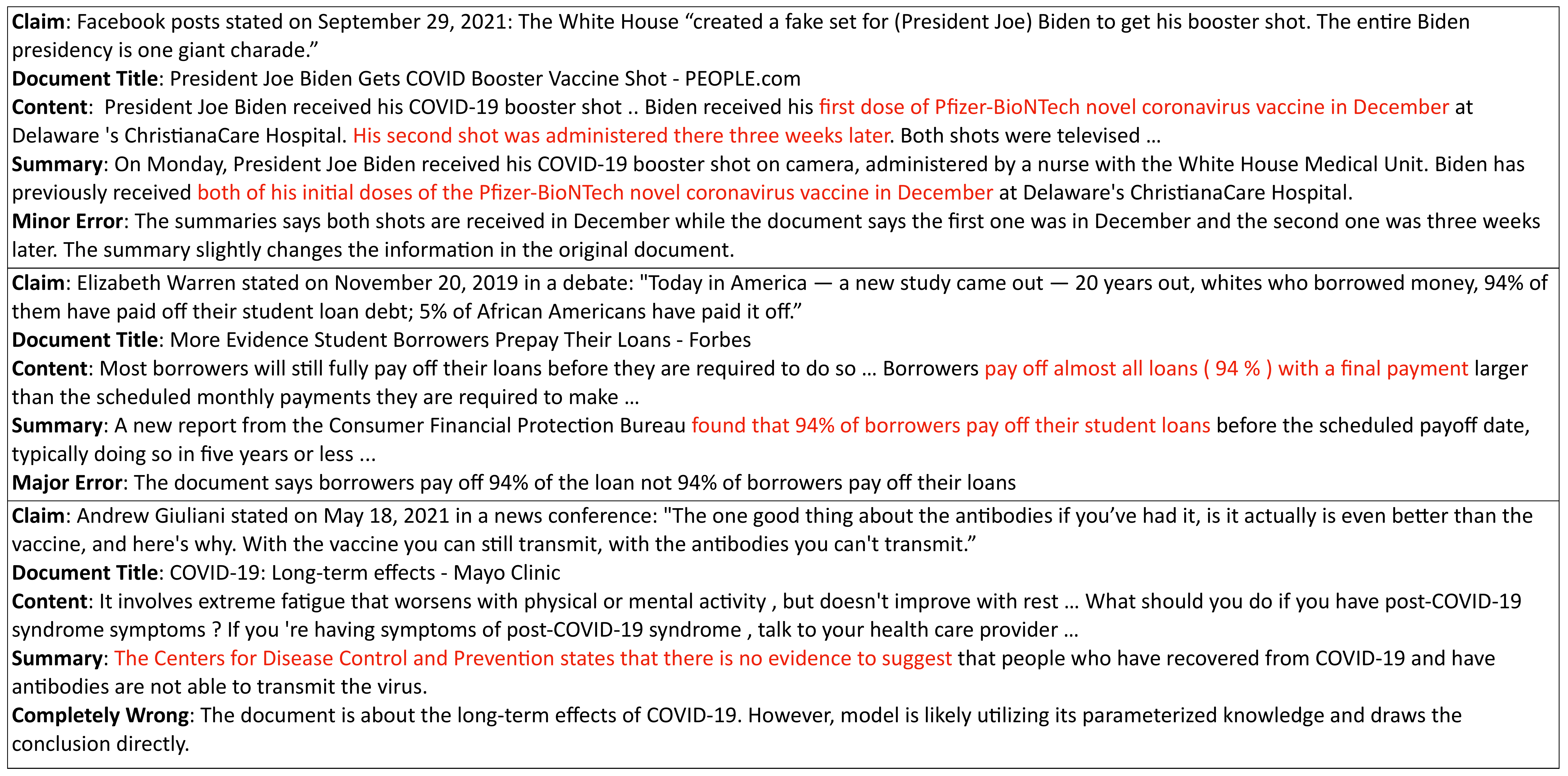}
  \caption{Three examples from the faithfulness evaluation (Section~\ref{sec:human-study-faithfulness}), showing the cases of minor error, major error, and completely wrong, respectively. Red text denotes the mismatches between the summary and the document.}
  \label{fig:unfaithful-examples}
\end{figure*}

Figure~\ref{fig:unfaithful-examples} shows three examples containing unfaithful content. We see that the ``Minor'' error does not affect the interpretation of the original document while ``Major'' and ``Completely Wrong'' errors alter the view.

% \begin{table*}
% \small
% \centering
% \begin{tabular}{cl cccc cccc}
% \toprule
%  \multicolumn{2}{c}{}   &         \multicolumn{4}{c}{Dev (N=200)} & \multicolumn{4}{c}{Test (N=200)} \\
% \textbf{Model} & \textbf{Evidence} & \textbf{Acc} & \textbf{Soft-Acc} & \textbf{Macro-F1} & \textbf{MAE} & \textbf{Acc} & \textbf{Soft-Acc} & \textbf{Macro-F1} & \textbf{MAE}\\ 
% \midrule
% \multirow{ 4}{*}{ChatGPT} & Claim only & 32.0 & 69.5 & 30.5 & 1.07 & 32.0 & 66.0 & 31.0 & 1.16\\
% & Claim + summary & 26.5 & 70.0 & 26.4 & 1.17 & 24.5 & 67.5 & 25.7 & 1.25\\

% & Claim + Gold subQs & 35.0 & 69.5 & 34.1 & 1.11 & 32.0 & 67.0 & 31.5 & 1.19\\
% & Claim + Gold subQs + Gold Answers & 40.0 & 80.0 & 40.5 & 0.85 & 38.0 & 77.0 & 35.8 & 0.96 \\

% % & Claim + subQs & 30.5 & 68.0 & 30.2 & 1.16 \\
% % \multirow{ 5}{*}{\shortstack{GPT \\ (prior knowledge restricted)}} & Claim + Human subclaims + Human answers* & 40.0 & 78.0 & 39.9 & 0.86\\
% % & Claim + Summary* & 29.5 & 73.0 & 28.1 & 1.08\\
% % & Claim + Summary + Gold subQs* & 26.0 & 73.5 & 25.7 & 1.13\\
% % & Claim + Summary + subQs* & 29.5 & 73.0 & 28.8 & 1.16\\
% % & Claim + Summary + Gold subQs + Gold answers* & 34.0 & 71.0 & 32.9 & 1.13\\
% \midrule
% \midrule

% \multirow{ 2}{*}{DeBERTa-large}  & Claim only & 37.0 & 71.0 & 34.6 & 0.98 & 25.5 & 68.0 & 27.5 & 1.12\\
% & Claim + summary & 52.5 & 88.5 & 54.5 & 0.64 & 57.5 & 93.0 & 57.8 & 0.50\\
% \bottomrule
% \end{tabular}
% \vspace{-0.2cm}
% \caption{Veracity classification performance on the test set of \textsc{ClaimDecomp} with different prompts using ChatGPT.}
% \label{tab:gpt-veracity-classification}
% \end{table*}
\subsection{Recruiting Process}

\begin{figure*}[t!]
  \centering
  \includegraphics[width=\textwidth]{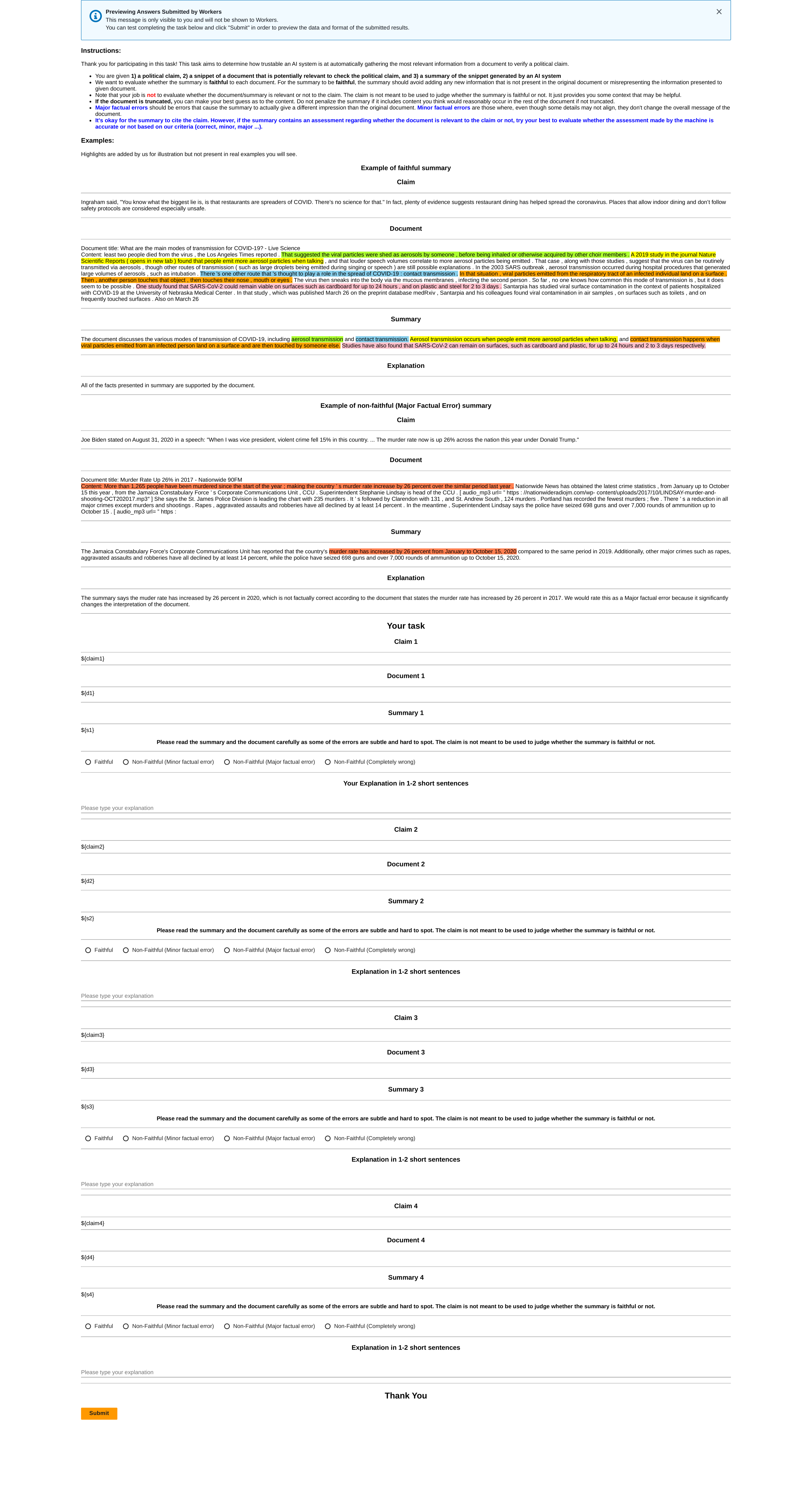}
  
  \caption{Interface of the faithfulness study we conducted in Section~\ref{sec:human-study-faithfulness}.}
  \label{fig:faithfulness_interface}
\end{figure*}

\paragraph{Faithfulness study}
We set up a qualification test that consists of 5 examples. We selected workers from MTurk if they get more than 3/5 examples correct according to our curated labels and if they write reasonable rationales. In total, there are 31 workers who took the qualification test and we selected 15 of them for the task. We pay \$3 for the qualification test and \$2 dollars for one HIT that contains 4 document-summary pairs in the actual task. The detailed instructions and the annotation interface is shown in Figure~\ref{fig:faithfulness_interface}.

\paragraph{Comprehensiveness study}
We set up a qualification test that consists of 10 examples. We selected workers from MTurk if they got more than 7/10 questions right according to our curated labels and if they write reasonable rationales. In total, there are 28 workers who took the qualification test and we selected 17 of them for the task. We pay \$3 for the qualification test and \$0.3 dollars for one question in the actual task.

\begin{figure*}[t!]
  \centering
  \includegraphics[width=\textwidth]{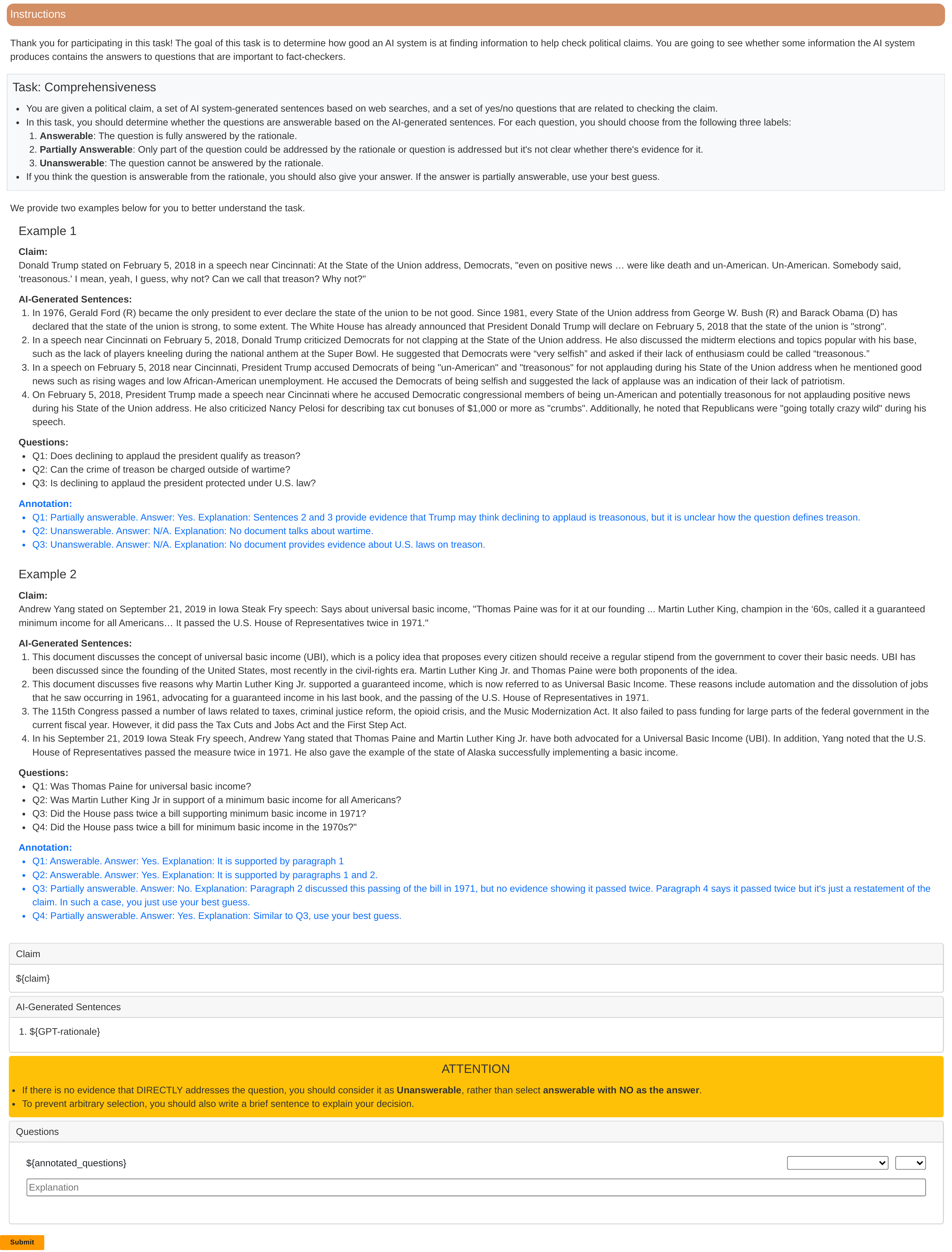}
  
  \caption{Interface of the comprehensiveness study we conducted in Section~\ref{sec:human-study-comprehensiveness}.}
  \label{fig:comprehensive_interface}
\end{figure*}

The detailed instructions and the annotation interface is shown in Figure~\ref{fig:comprehensive_interface}.
\begin{figure}[t!]
  \centering
  \includegraphics[width=0.5\textwidth]
  {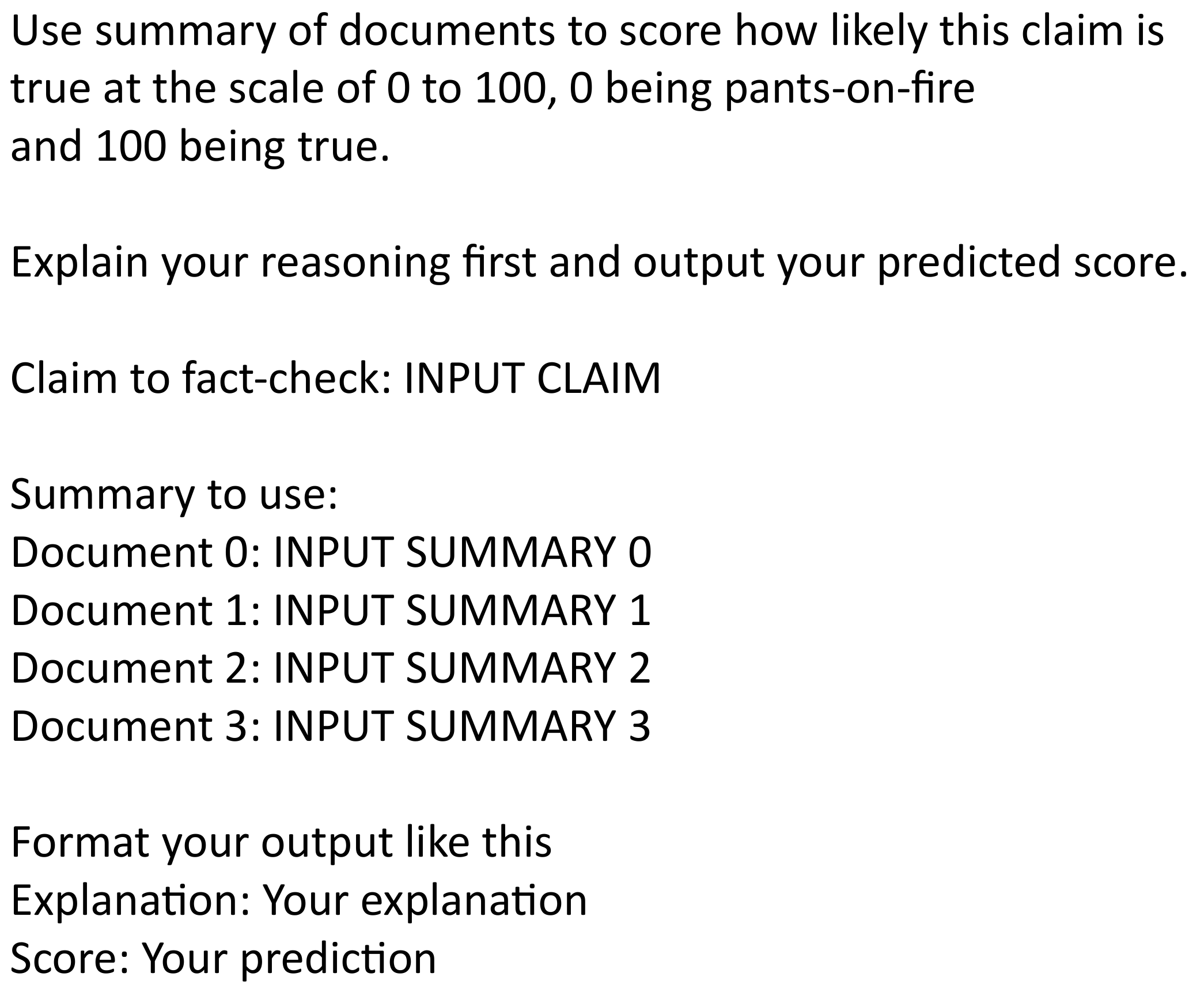}
  \caption{Zero-shot prompt for Claim + summary}
  \label{fig:score-producing-prompt}
\end{figure}

\begin{table*}
\small
\centering
\begin{tabular}{cl cccc}
\toprule
\textbf{Model} & \textbf{Evidence} & \textbf{Acc} & \textbf{Soft-Acc} & \textbf{Macro-F1} & \textbf{MAE}\\ 
\midrule
\multirow{2}{*}{ChatGPT} & Claim only & 32.0 & 66.0 & 31.0 & 1.16\\
                & Claim + summary  & 24.5 & 67.5 & 25.7 & 1.25\\

% & Claim + Gold subQs & 35.0 & 69.5 & 34.1 & 1.11 & 32.0 & 67.0 & 31.5 & 1.19\\
% & Claim + Gold subQs + Gold Answers & 40.0 & 80.0 & 40.5 & 0.85 & 38.0 & 77.0 & 35.8 & 0.96 \\
\midrule
\midrule

\multirow{ 2}{*}{DeBERTa-large}  & Claim only & 25.5 & 68.0 & 27.5 & 1.12\\
& Claim + summary  & 33.0 & 74.5 & 34.5 & 0.99 \\
\bottomrule
\end{tabular}
\caption{Veracity classification performance on the test set of \textsc{ClaimDecomp} with different prompts using ChatGPT.}
\label{tab:gpt-veracity-classification}
\end{table*}

\section{Using LLMs as a Veracity Classifier} \label{appendix:GPT-classifier}

We experiment with using ChatGPT (\texttt{gpt-3.5-turbo}) as the classifier in the final stage. Since ChatGPT is not trained on our training set, it does not have access to the label distribution of the dataset. To make a fair comparison with the DeBERTa model, instead of directly predicting a discrete label (one out of the six labels), we prompt the model to explain its reasoning process and predict a truthfulness score on a scale of 0 to 100, 0 for the claim being false and 100 for true. We then rank the examples according to the predicted scores and map the scores to discrete labels to the label distribution of the training set. To be specific, we rank the examples in the training set by their labels, assigning the lowest rank to \emph{pants-on-fire} and the highest to \emph{true}. Each label, denoted as $l_i$, corresponds to a percentile $p_i$. We then map the predicted score falling between $p_i$ and $p_{i+1}$ to the label $l_i$. We use a zero-shot prompt\footnote{We also experimented with few-shot prompts. However, these did not yield better performance than the zero-shot prompt.} to produce the score and the prompt is shown in Figure~\ref{fig:score-producing-prompt}.

% In addition to \textbf{claim only} and \textbf{Claim + summary}, which are used in the main experiments. We also experiment with two alternatives using ChatGPT. 
% \begin{itemize}
%     \item \textbf{Claim + Gold subQs}: We append the human-annotated subquestions~\citep{chen-etal-2022-generating} to the claim to see if they are helpful for ChatGPT to make a decision.
%     \item \textbf{Claim + Gold subQs + Answers}: We append the human-annotated subquestions~\citep{chen-etal-2022-generating} and the answers to the claim to see if they are helpful for ChatGPT to make a decision.
% \end{itemize}

The results are shown in Table~\ref{tab:gpt-veracity-classification}. Comparing the \textbf{claim-only} results from the two models, we see that ChatGPT achieves slightly better performance than DeBERTa. However, unlike the DeBERTa model, when adding the summary, we see a notable performance drop for ChatGPT. We argue that this might be because ChatGPT relies heavily on prior knowledge and it is not able to use the provided summary effectively. We believe improving this is a promising direction for future work.

% We ablate evidence in a prompt and test on the dev set of \textsc{ClaimDecomp} with both temporal and site retrieval constraints. We report the result in Table ~\ref{tab:gpt-veracity-classification}. We ask the model to explain about its reasoning process and predict the score at the scale of 0 to 100, 0 for claim being pants-on-fire and 100 for true. We transform the model's predicted score to calibrated thresholds for discrete labels to match the label distribution of the dataset. For example, there are 16 pants-on-fire on the dev dataset. We set the threshold for pants-on-fire as the 16th score after sorting the scores in ascending order. The GPT often already knows about the claim and makes the decision based on its knowledge. For the prompts, where the model is given enough evidence to make the correct judgment, we explicitly state the model should not use its prior knowledge about the claim when making judgement about the claim. We see that adding subclaims to the prompt generally helps with the model's performance compared to Claim only prompt using GPT model. Using GPT usually performs worse in all evaluation metrics in comparison to using DeBERTa-large. However, the GPT model can provide the explanation that humans can interpret. We believe that for the human-in-the-loop fact-checking system, GPT could be better suit.

\end{document}